%% file: main.tex
\pdfoutput=1

\documentclass[11pt]{article}

\usepackage{times}
\usepackage{latexsym}

\usepackage{authblk}

\usepackage[T1]{fontenc}

\usepackage[utf8]{inputenc}

\usepackage{microtype}
\usepackage{acl}
\usepackage{caption}
\usepackage{subcaption}

\definecolor{googleblue}{HTML}{4285F4}
\definecolor{googlered}{HTML}{DB4437}
\definecolor{googlepurple}{HTML}{A142F4}
\definecolor{googlegreen}{HTML}{0F9D58}

\input{preamble}

\usepackage{nicefrac}
\usepackage{algorithm}
\usepackage[noend]{algpseudocode}
\usepackage{float}

\DeclareMathAlphabet{\mathbcal}{OMS}{cmsy}{b}{n}
\usepackage{amsmath}
\usepackage{mathrsfs}
\usepackage{comment}
\usepackage{bm}
\usepackage{bbm}
\usepackage{amssymb}
\usepackage{paralist}
\usepackage{xcolor}
\usepackage{graphicx}

\usepackage{inconsolata}

\usepackage[strict]{changepage}

\usepackage{framed}

\usepackage{natbib}

\definecolor{formalshade}{rgb}{0.95,0.95,1}

\newenvironment{formal}{
  
  \MakeFramed{\advance\hsize-\width\FrameRestore}
  \noindent\hspace{-4.55pt}
  \begin{adjustwidth}{4pt}{7pt}
  
}
{
  \end{adjustwidth}\endMakeFramed
}

\usepackage{colortbl}

\usepackage{afterpage}

\author[1]{Jiangnan Fang}
\author[1]{Cheng-Tse Liu}
\author[1]{Jieun Kim}
\author[1]{Yash Bhedaru}
\author[1]{Ethan Liu}
\author[1]{Nikhil Singh}

\author[2]{\\ Nedim Lipka}
\author[2]{Puneet Mathur}
\author[2]{Nesreen K. Ahmed}
\author[2]{Franck Dernoncourt}
\author[2]{\\ Ryan A. Rossi}
\author[2]{Hanieh Deilamsalehy}

\affil[1]{University of California, Santa Cruz}
\affil[2]{Adobe Research}

\title{Multi-LLM Text Summarization}

\begin{document}

\newpage
\maketitle

\begin{abstract}
In this work, we propose a Multi-LLM summarization framework, and investigate two different multi-LLM strategies including centralized and decentralized. Our multi-LLM summarization framework has two fundamentally important steps at each round of conversation: generation and evaluation. These steps are different depending on whether our multi-LLM decentralized summarization is used or centralized. In both our multi-LLM decentralized and centralized strategies, we have $k$ different LLMs that generate diverse summaries of the text. However, during evaluation, our multi-LLM centralized summarization approach leverages a single LLM to evaluate the summaries and select the best one whereas $k$ LLMs are used for decentralized multi-LLM summarization. Overall, we find that our multi-LLM summarization approaches significantly outperform the baselines that leverage only a single LLM by up to 3x. These results indicate the effectiveness of multi-LLM approaches for summarization.
\end{abstract}

\section{Introduction}

Large language models (LLMs) have been shown to have the potential to produce high-quality summaries \cite{{chowdhery2022palmscalinglanguagemodeling}, {zhang2023benchmarking}, {goyal2023news}, {pu2023summarization}}. However, despite the remarkable progress in LLM-based summarization, limitations still exist for documents where useful information may be sparsely distributed throughout the text. Research by \cite{liu2023lost} highlights that a naive application of LLMs may overlook critical details or fail to grasp the holistic meaning of a document, indicating the need for more refined methods.

To address this, recent efforts have explored prompt-engineering techniques to guide LLMs towards producing better summaries \cite{adams2023sparsedensegpt4summarization}. These techniques, while promising, still face limitations in consistently delivering high-quality summaries across different document types and structures. Instead of relying solely on a single model or simple prompt-engineering methods, we propose an approach novel to the summarization domain that focuses on aggregating the collective strengths of multiple LLMs. By combining the capabilities of multiple models with a diverse set of knowledge bases, we show it's possible to achieve more robust summaries across domains.

\medskip\noindent\textbf{Summary of Main Contributions.} 
The main contributions of this work are as follows:

\begin{compactitem}%
    \item We propose the first framework for multi-LLM text summarization and investigate two topologies: centralized and decentralized.

    \item We find that multi-LLM text summarization often performs better than using a single LLM for summarization, and we show that the best performing method in the framework aligns with human judgments.

    \item We conduct experiments on how prompting, number of LLMs, and various combinations of generating and evaluating LLMs can affect quality of summaries in the multi-LLM setup.

\end{compactitem}

\section{Related Work} \label{sec:related-work}

\subsection{Summarization}
Recent advancements in summarization have increasingly leveraged large language models (LLMs), moving beyond fine-tuned transformer models like Pegasus, BART, and T5. Studies consistently show that LLMs can generate summaries with higher coherence, relevance, and factual accuracy, often rivaling or surpassing human-written summaries \cite{goyal2023news, zhang2023benchmarking, pu2023summarization}.

For example, \citet{goyal2023news} demonstrated that GPT-3 (text-davinci-002) produced summaries preferred by human evaluators over fine-tuned models like Pegasus and BRIO on structured datasets such as CNN/DM \cite{nallapati2016abstractivetextsummarizationusing} and XSUM \cite{narayan2018dontdetailsjustsummary}. Similarly, \citet{zhang2023benchmarking} emphasized the importance of instruction tuning in achieving superior zero-shot performance for summarization tasks. \citet{pu2023summarization} further highlighted improved factual consistency and reduced hallucinations when using LLMs.

While these studies validate the potential of LLMs in summarizing well-structured texts, they may falter for inputs lacking clear structural cues and exhibiting greater complexity. Research focusing on longer text summarization, such as \citet{keswani2024abstractive}, employed semantic clustering and multi-stage summarization with LLaMA2 to manage lengthy inputs. However, such approaches often rely on predefined hierarchical processing strategies that may oversimplify the nuanced relationships within the text. Moreover, as \citet{liu2023lost} noted, LLMs tend to neglect content from the middle sections of longer documents, resulting in incomplete or unbalanced summaries.

Our work aims to improve performance for both long and short text summarization, and it builds upon aforementioned foundations by proposing a multi-LLM framework designed to overcome these shortcomings through information exchange and collaborative synthesis.

\subsection{Multi-LLM}

The concept of leveraging multiple LLMs collaboratively has gained traction in recent research, particularly for tasks requiring complex reasoning and factual accuracy. For instance, \citet{liang2024encouragingdivergentthinkinglarge} introduced the Multi-Agent-Debate (MAD) framework, where LLMs engage in iterative debates to refine their reasoning. This framework demonstrated that a multi-agent GPT-3.5-Turbo setup outperformed GPT-4 on reasoning datasets. Similarly, \citet{chen2024reconcileroundtableconferenceimproves} proposed RECONCILE, a framework where LLMs collaboratively refine answers and explanations, achieving significant improvements over single-agent systems. \citet{li2024improvingmultiagentdebatesparse} extended this line of research by optimizing agent connections, showing that sparse networks can maintain performance while reducing computational overhead.

Although these studies reveal the potential of multi-LLM approaches, their focus remains on structured reasoning tasks, such as question answering and fact-checking. They have not been adequately explored in the context of synthesizing distributed information, addressing content imbalances, and preserving the coherence of summaries across extended texts.

We hope to bridge this gap by adapting multi-LLM frameworks to the domain of document summarization, addressing limitations of both single LLM and traditional hierarchical techniques, and positioning multi-LLM summarization as a promising solution.

\definecolor{categoryblue}{HTML}{4285F4}
\definecolor{categoryred}{HTML}{DB4437}
\definecolor{categorypurple}{HTML}{A142F4} 
\definecolor{categorygreen}{HTML}{0F9D58}

\begin{table*}[h]
\centering
\small
\begin{tabular}{c rc}
\toprule
\textbf{Multi-LLM Summarization} & \\
\textbf{Framework} & \textbf{General Mechanism} & \textbf{Stage}  \\ 
\midrule

\multirow{4}{*}{{\textsc{\textbf{Centralized}}} (\textbf{Sec.~\ref{sec:cent-multi-llm-summarization}})} 
& 
\multirow{2}{*}{{\text{{Single-Round}}} (\text{Sec.~\ref{sec:cent-single-round}})} 
& \text{Generation} (\S~\ref{sec:cent-single-round-gen}) \\ 
& & \text{Evaluation} (\S~\ref{sec:cent-single-round-eval}) \\
\cmidrule{2-3}
& 
\multirow{2}{*}{{\text{{Conversational}}} (\text{Sec.~\ref{sec:cent-convo}})} 
& \text{Generation} (\S~\ref{sec:cent-convo-gen}) \\ 
& & \text{Evaluation} (\S~\ref{sec:cent-convo-eval}) \\
\midrule

\multirow{4}{*}{{\textsc{\textbf{Decentralized}}} (\textbf{Sec.~\ref{sec:decent-multi-llm-summarization}})} 
& 
\multirow{2}{*}{{\text{{Single-Round}}} (\text{Sec.~\ref{sec:decent-single-round}})} 
& \text{Generation} (\S~\ref{sec:decent-single-round-gen}) \\ 
& & \text{Evaluation} (\S~\ref{sec:decent-single-round-eval}) \\
\cmidrule{2-3}
& 
\multirow{2}{*}{{\text{{Conversational}}} (\text{Sec.~\ref{sec:decent-convo}})} 
& \text{Generation} (\S~\ref{sec:decent-convo-gen}) \\ 
& & \text{Evaluation} (\S~\ref{sec:decent-convo-eval}) \\

\bottomrule
\end{tabular}
\caption{
Overview of Multi-LLM Summarization Framework (Sections~\ref{sec:cent-multi-llm-summarization}-\ref{sec:decent-multi-llm-summarization}).
}
\label{tab:multi-llm-summarization-overview}
\end{table*}

\section{Multi-LLM Summarization Framework} \label{sec:framework}

\begin{figure}[h!]

    \centering
    \begin{subfigure}[b]{0.45\linewidth}
        \centering
        \includegraphics[width=\linewidth]{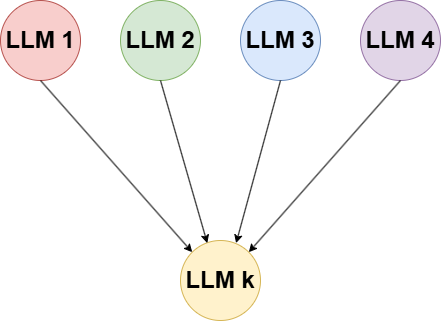}
        \caption{Centralized}
        \label{fig:centralized}
    \end{subfigure}
    \hfill
    \begin{subfigure}[b]{0.45\linewidth}
        \centering
        \includegraphics[width=\linewidth]{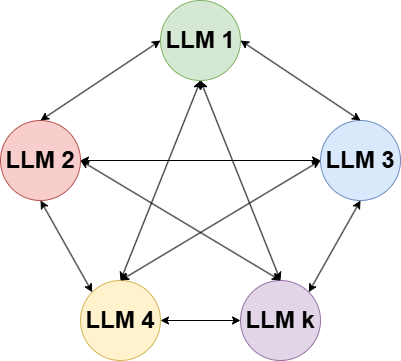}
        \caption{Decentralized}
        \label{fig:decentralized}
    \end{subfigure}
    \caption{%
    Centralized and Decentralized approaches using a 5-LLM example. Similar topologies can be applied to any ("$k$") number of LLMs. In centralized interactions, all models communicate with a central model; in decentralized interactions, each model communicate with every other model and also itself.
    }
    \label{fig:centralized_vs_decentralized}

\end{figure}

In this work, we propose a novel multi-LLM summarization framework that leverages multiple large language models to enhance summarization quality of long document input. Through the distribution of generation and evaluation of candidate summaries across multiple models, our framework aims to provide better summaries than single LLM methods, leveraging expertise from different models. We present two interaction topologies, \textbf{centralized} and \textbf{decentralized}, to guide the collaboration, evaluation, and refinement of summaries between LLMs. Visually these two methods can be represented at a high level in Figure~\ref{fig:centralized_vs_decentralized}. In the datasets we test, articles are typically tens of thousands of words long and exceed the context window of most standard LLMs. To handle this, we establish a two stage process that involves chunking the source document, independently summarizing each chunk of the source document, and then applying a second round of chunking and summarization on the concatenated intermediate results. Throughout both these stages, both frameworks allow multiple LLMs to collaborate and converge on a single final high quality summary of the entire original reference document. Table~\ref{tab:multi-llm-summarization-overview} provides an overview of our framework's four main variations.

\section{Centralized Multi-LLM Summarization} \label{sec:cent-multi-llm-summarization}

The steps for centralized summarization can be found in Algorithm~\ref{alg:multi-llm-hier-summary-centralized}. This method leverages multiple LLMs to generate candidate summaries and uses a central LLM to evaluate their quality and guide iterative refinements.

\subsection{Single Round} \label{sec:cent-single-round}

In the simplest case, we prompt each LLM once, gather their summaries, and then perform a single evaluation step to select the best final summary. This is the initial process before we extend it to multiple rounds.

\subsubsection{Generation Phase} \label{sec:cent-single-round-gen}

In the single-round setting, each LLM from the list of participating models $\mathcal{M} = \{M_1, \dots, M_k\}$ independently generates a summary of the same input text using a common prompt $P$. The prompt $P$ is illustrated in Figure~\ref{fig:initial-prompt-for-generating-summary}. Formally, for each LLM $M_j \in \mathcal{M}$, the output is
$$S_j = M_j(P, S)$$

where $S$ represents the input text. Running this step for all $M_j$ yields a set of summaries $\mathcal{S} = \{S_1, \dots, S_k\}$.

This initial generation stage corresponds to line~\ref{algline:hier-summary-gen-summary-cent}
of Algorithm~\ref{alg:multi-llm-hier-summary-centralized}. Conceptually, each model contributes its unique perspective, leading to a diverse pool of candidate summaries, which is important for robust summary selection in the following evaluation phase.

\subsubsection{Evaluation Phase} \label{sec:cent-single-round-eval}

After collecting the set of candidate summaries $\mathcal{S}$, we select a central agent $C \in \mathcal{M}$ to evaluate these summaries. The central LLM $C$ uses an evaluation prompt $P_{ec}$, as shown in Figure~\ref{fig:prompt-for-evaluating-summary-centralized}, to assess the quality of each summary. To reduce potential bias arising from authorship attribution, we use anonymized identifiers for summaries like \texttt{agent\_1}, \texttt{agent\_2}, etc. during evaluation.

Formally, we obtain $E = C(P_{ec}, \mathcal{S})$, where $E$ is the central LLM's evaluation of all candidate summaries. This includes the choice for the best summary (expressed as its anonymized identifier) and a confidence score for that evaluation (expressed as an integer from 0 to 10), denoted together as $\vr = \textsc{AggrResults}(E)$ in Algorithm~\ref{alg:multi-llm-hier-summary-centralized}. We de-anonymize the identifier to recover the text of the selected summary $S_j$ and set this as our final output $S^{*}$. In the single-round regime, this terminates the process as no further iterations are performed.

In the evaluation prompt, we include the prompt to output a confidence score so there is a variable on which to impose a stopping condition. This allows us to extend the centralized process to multiple rounds of generation and evaluation using that condition. This process is explained in subsequent sections.

\subsection{Conversational} \label{sec:cent-convo}

In the conversational approach, we repeat the generation and evaluation phases multiple times. We define each generation-evaluation process as one round and define conditions under which the process ends or a new round should begin, up to a maximum number of rounds.

\subsubsection{Generation Phase} \label{sec:cent-convo-gen}

The first round of the conversational approach mirrors the single-round procedure (Section~\ref{sec:cent-single-round-gen}). Each LLM $M_j$ generates an initial summary $S_j^{(1)}$ from the original input text $S$ using the prompt $P$:
\[
S_j^{(1)} = M_j(P, S).
\]

If the evaluation result from the previous round has a confidence score less than the threshold or, if the LLM fails to output a readable confidence score, the pipeline proceeds to the next round. For the second and subsequent rounds, we use the prompt $P^{(i)}$, shown in Figure~\ref{fig:prompt-for-generating-better-summary-via-convo-multi-LLM}. LLMs in the second and subsequent rounds have access to both the text to be summarized and summaries from the previous round. Concretely, in round $i > 1$:
\[
S_j^{(i)} = M_j(P^{(i)}, S).
\]

The hope is that LLM is able to iteratively improve summarization based upon previous outputs from itself and other models.

\subsubsection{Evaluation Phase} \label{sec:cent-convo-eval}

The evaluation phase in round $i > 1$ is conceptually similar to the single-round setting (Section~\ref{sec:cent-single-round-eval}), but now operates on candidate summaries generated immediately before in the generation phase $\mathcal{S}_i = \{S_1^{(i)}, \dots, S_k^{(i)}\}$. The central LLM $C$ evaluates these candidates using $P_{ec}$:
\[
E^{(i)} = C(P_{ec}, \mathcal{S}_i),
\]

If the confidence level meets the threshold, the process terminates, and the summary chosen by the central LLM is accepted as $S^{*}$. Otherwise, we proceed to the next round of summary generation and evaluation. For the confidence scores we have chosen the range 0-10 as it is fine-grained but also is one of the most common rating scales.

\subsection{Analysis of Complexity}
The centralized approach uses $k$ models for generation and 1 central model for evaluation; other than text length, the number of input tokens scale linearly with the number of models and with the number of rounds. Output tokens also scale linearly with number of models and number of rounds, but since we instruct the model to output a fixed number of words for summary (and in our experiments the models are largely compliant), and output only the anonymous identifier for a chosen summary, we ensure bounded runtime and cost. Further analysis can be found at Appendix~\ref{sec:cent-analysis-and-discussion}.

\begin{algorithm}[t!]
\small
\caption{
Centralized Multi-LLM 
Summary
}
\label{alg:multi-llm-hier-summary-centralized}
\begin{algorithmic}[1]
\Require
ordered set $\mathcal{S} = \{S_1,\ldots,S_m\}$ of summaries,
set $\mathcal{M} = \{M_1,\ldots,M_k\}$ of $k$ LLMs,
a central agent $C \in \mathcal{M}$, 
max number of conversational rounds $t_{\max}$,
initial summarization prompt $P$ (\eg, Figure~\ref{fig:initial-prompt-for-generating-summary}),
evaluation prompt $P_ec$ (\eg, Figure~\ref{fig:prompt-for-evaluating-summary-centralized}) for centralized version 

\Ensure summary $S^{*}$ of the text 

\State $S = \textsc{CreateSummary}(\mathcal{S})$ 
\label{algline:create-summary-cent}

\For{$i=1$ to $t_{\max}$} \Comment{conversation rounds} \label{algline:hier-summary-rounds-of-conversation-cent}

\For{\textbf{each} model $M_j \in \mathcal{M}$} \label{algline:hier-summary-foreach-model-gen-step-cent}
    \State $S_{j}^{(i)} = M_j(P, S)$ \label{algline:hier-summary-gen-summary-cent}
\EndFor
\State Let $\mathcal{S}_{i} = \{S_{1}^{(i)}, S_{2}^{(i)}, \ldots, S_{k}^{(i)}\}$ \label{algline:hier-summary-let-S-cent}

\State $E^{(i)} = C(P_{ec}, \mathcal{S}_{i})$

\State $\vr = \textsc{AggrResults}(E^{(i)})$

\State $j \leftarrow \argmax_{M_j \in \mathcal{M}} r_j$
\State Set $S^{*} \leftarrow S_{j}^{(i)}$
\If{$\textsc{Converged}(\vr)$} 
    \textbf{return} $S^{*}$
\EndIf
\State Set $P$ to prompt in Figure~\ref{fig:prompt-for-generating-better-summary-via-convo-multi-LLM}.
\EndFor
\end{algorithmic}
\end{algorithm}

\begin{figure}[t!]
\begin{formal}
\small
\textit{\tt 
\\
Provide a concise summary of the text in around 160 words. Output the summary text only and nothing else.
\begin{center}
[text]
\end{center}
}
\end{formal}
\caption{%
Prompt for generating the initial summary in the first round.
}
\label{fig:initial-prompt-for-generating-summary}
\end{figure}

\begin{figure}[t!]
\begin{formal}
\small
\textit{\tt 
\\
Given the original text below, along with the summaries of that text by [k] LLMs, please generate a better summary of the original text in about 160 words.\\
\\
ORIGINAL: 
\begin{center}
[text]
\end{center} 
Summary by $M_1$:
\begin{center}
[LLM 1's summary]\\
$\vdots$
\end{center}
Summary by $M_k$:
\begin{center}
[LLM k's summary]
\end{center}
}
\end{formal}
\caption{%
Generation prompt that is used after the initial round of conversation among the multiple LLMs. 
Note that the above prompt is for generating the final summary, however, for the chunk-level generation, it would just be the actual chunk.
}
\label{fig:prompt-for-generating-better-summary-via-convo-multi-LLM}
\end{figure}

\begin{figure}[t!]
\begin{formal}
\small
\textit{\tt 
\\
Given the original text below, along with the summaries of that text by [k] agents, please evaluate the summaries and output the name of the agent that has the best summary. Output the exact name only and nothing else.\\
\\
ORIGINAL:
\begin{center}
[chunk or concatenated chunk summaries S] 
\end{center} 
Summary by agent\_1:
\begin{center}
[LLM 1's summary]\\
$\vdots$
\end{center}
Summary by agent\_k:
\begin{center}
[LLM k's summary]
\end{center}
}
\end{formal}
\caption{%
Evaluation prompt for evaluating the summaries generated by different LLMs using our conversational (decentralized) multi-LLM framework. "k" is a parameter reflecting the number of LLMs that generate summaries.
}
\label{fig:prompt-for-evaluating-summary-decentralized}
\end{figure}

\begin{figure}[t!]
\begin{formal}
\small
\textit{\tt 
\\
Given the initial text below, along with the summaries of that text by [k] LLMs, please evaluate the generated summaries and output the name of the LLM has the best summary. On a separate line indicate a confidence level between 0 and 10.
\\
ORIGINAL:
\begin{center}
[text]
\end{center} 
Summary by $M_1$:
\begin{center}
[LLM 1's summary]\\
$\vdots$
\end{center}
Summary by $M_k$:
\begin{center}
[LLM k's summary]\\
\end{center}
Remember, on a separate line indicate a confidence level between 0 and 10
}
\end{formal}
\caption{%
Evaluation prompt for evaluating the summaries generated using our conversational (centralized) multi-LLM framework.
More specifically, we have added an instruction for centralized multi-LLM summarization approach that in addition to providing the best summary, it also outputs the confidence level between 0 and 10. "k" is a parameter reflecting the number of summary-generating LLMs.
}
\label{fig:prompt-for-evaluating-summary-centralized}
\end{figure}

\begin{figure}[htbp]
\begin{formal}
\small
\textit{\tt 
\\
Provide a concise summary of the text in around 160 words. Output the summary text only and nothing else.
\begin{center}
[concatenated chunk summaries S]   
\end{center} 
}
\end{formal}
\caption{%
Generation prompt for generating the final summary from the summarized chunks using our conversational (decentralized) multi-LLM framework. This prompt is the same as the one for the initial summary.
}
\label{fig:prompt-for-generating-final-summary}
\end{figure}

\section{\!\!\!\!Decentralized Multi-LLM Summarization}  \label{sec:decent-multi-llm-summarization}

Previously we introduced the summarization procedure for centralized approach (Section~\ref{sec:cent-multi-llm-summarization}), which diversifies the knowledge base for summarization. We extend the paradigm for the evaluator as well. In the decentralized approach, multiple LLMs also participate in the evaluation process with the hope that a best summary decided on consensus is more robust compared to a single model's decision.

\subsection{Single Round} \label{sec:decent-single-round}

\subsubsection{Generation Phase} \label{sec:decent-single-round-gen}
Generation procedure is the same as that in the centralized approach described in Section~\ref{sec:cent-single-round-gen}. As before, multiple LLMs independently generate summaries for the input text, obtaining the list of summaries $\mathcal{S} = \{S_1, \ldots, S_k\}$.

\subsubsection{Evaluation Phase} \label{sec:decent-single-round-eval}

For evaluation, each model that authored a summary is prompted with a new evaluation prompt (Figure~\ref{fig:prompt-for-evaluating-summary-decentralized}) which does not include a confidence level and receives the text to be summarized along with summaries authored by all agents including itself. More formally, model preferences $E_{1}^{(i)}, \ldots, E_{k}^{(i)}$ are collected, where each $E_{j}^{(i)}$ represents model $M_j$'s choice of the best summary among ${S_{1}^{(i)}, \ldots, S_{k}^{(i)}}$. These preferences are aggregated into a result vector $\vr \in {1,\ldots,k}^k$, where each element $r_j$ indicates which model's summary was chosen by model $M_j$. Convergence is achieved when a majority of models select the same summary, formally expressed as $\exists m \in {1,\ldots,k} : |{j : r_j = m}| > \frac{k}{2}$. \footnote{Here our implementation requires votes exceeding absolute majority for a summary to be immediately selected. In the case of 2 LLMs, this is equivalent to a unanimous decision because one vote does not satisfy absolute majority.}
When no majority choice emerges, the single-round approach ($t_{\max} = 1$) the algorithm selects the summary from a designated tie-breaker model $M_t$, where $t \in {1,\ldots,k}$. Since the tie-breaker model can be any model in the multi-LLM setup, we run experiments with different choices of evaluator and tie-breaking models. Formally, the final summary $S^*$ is determined as:
$$ S^* = 
\begin{cases}
S_m^{(1)} & \text{if } \exists m : |\{j : E_{j}^{(1)} = m\}| > \frac{k}{2} \\
S_t^{(1)} & \text{if } \max_{l} |\{j : E_{j}^{(1)} = l\}| \leq \frac{k}{2}
\end{cases}
$$
where $m \in {1,\ldots,k} : |{j : r_j = m}| > \frac{k}{2}$.
We test the different choices of tie-breaker model in the experiment Appendix~\ref{sec:exp-varying-centralized-llm}.

\subsection{Conversational} \label{sec:decent-convo}

The conversational approach extends the decentralized framework by introducing multiple rounds of generation and evaluation phases. Each generation-evaluation cycle constitutes a round, with iterations continuing until either consensus is achieved or a maximum number of rounds ($t_{\max}$) is reached.

\subsubsection{Generation Phase} \label{sec:decent-convo-gen}

Generation follows the methodology in Section~\ref{sec:cent-single-round-gen}, producing the set of summaries $\mathcal{S} = {S_1, \ldots, S_k}$. A key distinction from the single-round approach lies in the conditional regeneration mechanism: when consensus fails in the first round, subsequent rounds use a new prompt (Figure~\ref{fig:prompt-for-generating-better-summary-via-convo-multi-LLM}) which includes generated summaries from previous evaluations.

\subsubsection{Evaluation Phase} \label{sec:decent-convo-eval}

The first round of evaluation is identical to that in the single-round approach, but enters additional rounds with new generation prompts. Formally, let $E_{j}^{(i)}$ represent model $M_j$'s choice in round $i$. In the single-round case, non-consensus (when $\max_{m} |\{j : E_{j}^{(i)} = m\}| \leq \frac{k}{2}$) triggers an immediate fallback to a tie-breaker model. In contrast, the conversational approach initiates a new generation-evaluation round with an updated prompt (Figure~\ref{fig:prompt-for-generating-better-summary-via-convo-multi-LLM}). This process continues until either a majority consensus emerges or $t_{\max}$ rounds are exhausted. After $t_{\max}$ rounds without a consensus, the algorithm defaults to the tie-breaker mechanism described in Section~\ref{sec:decent-single-round-eval}.

\subsection{Analysis of Complexity}
The decentralized approach uses $k$ models for both generation and evaluation. For this reason the input and output tokens scale quadratically with number of models. As before, we instruct the model to output a fixed number of words for summary and an identifier only for evaluation and so ensure bounded runtime and cost. Further analysis can be found at Appendix~\ref{sec:decent-analysis-and-discussion}.

\begin{algorithm}[t!]
\small
\caption{
Decentralized Multi-LLM 
Summary
}
\label{alg:multi-llm-hier-summary-decentralized}
\begin{algorithmic}[1]
\Require
ordered set $\mathcal{S} = \{S_1,\ldots,S_m\}$ of summaries,
set $\mathcal{M} = \{M_1,\ldots,M_k\}$ of $k$ LLMs,
max number of conversational rounds $t_{\max}$,
initial summarization prompt $P$ (\eg, Figure~\ref{fig:initial-prompt-for-generating-summary}),
evaluation prompt $P_e$ (\eg, Figure~\ref{fig:prompt-for-evaluating-summary-decentralized})
\Ensure summary $S^{*}$ of the text 

\State $S = \textsc{CreateSummary}(\mathcal{S})$ 
\label{algline:create-summary-decent}

\For{$i=1$ to $t_{\max}$} \Comment{conversation rounds} \label{algline:hier-summary-rounds-of-conversation-decent}

\For{\textbf{each} model $M_j \in \mathcal{M}$} \label{algline:hier-summary-foreach-model-gen-step-decent}
    \State $S_{j}^{(i)} = M_j(P, S)$ \label{algline:hier-summary-gen-summary-decent}
\EndFor
\State Let $\mathcal{S}_{i} = \{S_{1}^{(i)}, S_{2}^{(i)}, \ldots, S_{k}^{(i)}\}$ \label{algline:hier-summary-let-S-decent}

\For{\textbf{each} model $M_j \in \mathcal{M}$ } \label{algline:hier-summary-foreach-model-eval-step}
    \State $E_{j}^{(i)} = M_j(P_e, S_{1}^{(i)}, \ldots, S_{k}^{(i)})$ \label{algline:hier-summary-eval-summary}
\EndFor

\State Set $\mathcal{E}_{i} = \{E_{1}^{(i)}, E_{2}^{(i)}, \ldots, E_{k}^{(i)}\}$ \label{algline:hier-summary-let-E}

\State $\vr = \textsc{AggrResults}(E_{1}^{(i)}, \ldots, E_{k}^{(i)})$
\State $j \leftarrow \argmax_{M_j \in \mathcal{M}} r_j$
\State Set $S^{*} \leftarrow S_{j}^{(i)}$
\If{$\textsc{Converged}(\vr)$} 
\textbf{return} $S^{*}$
\EndIf

\State Set $P$ to prompt in Figure~\ref{fig:prompt-for-generating-better-summary-via-convo-multi-LLM}.
\EndFor

\end{algorithmic}
\end{algorithm}

\section{Experiments} \label{sec:exp}
To investigate the proposed multi-LLM summarization framework, we conduct extensive experiments to evaluate its effectiveness.

\subsection{Experimental Setup}
We use ArXiv \cite{cohan2018discourseawareattentionmodelabstractive} and GovReport \cite{huang-etal-2021-efficient} to evaluate our summarization methods. We assess the quality of LLM-generated summaries using ROUGE-1, ROUGE-L, BLEU-1, and BLEU-4 metrics. 
For comparison with our multi-LLM approach, unless otherwise mentioned, we leverage GPT-3.5, GPT-4o, GPT-4o mini, and LLaMA3-8B as baselines.
For these models, we perform the same chunking across all models, and the summarization prompt is identical to that in the first round of the multi-LLM process (Figure~\ref{fig:prompt-for-generating-final-summary}).
Unless otherwise mentioned, all models use 4K-char chunk-size, and the final summary represents a concatenation of the generated summaries. Finally, unless otherwise mentioned, we set $W=160$ for all the models.

\begin{table*}[htp]
\resizebox{1.0\textwidth}{!}{
\begin{tabular}{ll ccHccH ccHccH}
\toprule
 & & \multicolumn{6}{c}{\bf{ArXiv}} & \multicolumn{6}{c}{\bf{GovReport}} \\
\cmidrule(lr){3-7}
\cmidrule(lr){8-14}
  & 
 & ROUGE-1 $\uparrow$ & ROUGE-L $\uparrow$ & METEOR & BLEU-1 $\uparrow$ & BLEU-4 $\uparrow$ & BERT Score $\uparrow$ & ROUGE-1 $\uparrow$ & ROUGE-L $\uparrow$ & METEOR $\uparrow$ & BLEU-1 $\uparrow$ & BLEU-4 $\uparrow$ & BERT Score $\uparrow$\\
 
\midrule
\multirow[t]{2}{*}{} & LLaMA3-8B & 0.180 & 0.106 & 0.263 & {0.084} & {0.021} & 0.0 & 0.403 & 0.177 & 0.377 & {0.242} & {0.079} & 0.0 \\
& GPT-3.5 & 0.193 & 0.114 & 0.275 & {0.093} & {0.026} & 0.827 & 0.390 & 0.178 & \textbf{0.383} & {0.226} & {0.084} & \textbf{0.857} \\
 & GPT-4o mini & 0.217 & 0.118 & 0.281 & {0.108} & {0.020} & 0.824 & 0.384 & 0.156 & 0.363 & {0.224} & {0.058} & 0.853 \\
 & GPT-4o & 0.165 & 0.095 & 0.248 & {0.073} & {0.015} & 0.823 & 0.372 & 0.155 & 0.364 & {0.211} & {0.059} & 0.854 \\
 
\midrule
\multirow{2}{*}{\bf Decentralized}
& Multi-LLM 3 round max & 0.313 & 0.163 & 0.301 & {0.200} & {0.029} & 0.829 & 0.447 & 0.180 & 0.292 & {0.458} & {0.098} & 0.855 \\
 & Multi-LLM 1 round max & \textbf{0.339} & \textbf{0.180} & \textbf{0.315} & {\textbf{0.224}} & {\textbf{0.043}} & \textbf{0.832} & 0.468 & 0.190 & 0.305 & {0.477} & {0.112} & \textbf{0.857} \\
\midrule
\multirow{2}{*}{\bf Centralized}
& Multi-LLM 3 round max & 0.329 & 0.168 & 0.302 & {0.217} & {0.031} & 0.830 & {0.468} & {0.189} & 0.307 & {0.470} & {0.109} & \textbf{0.857} \\
 & Multi-LLM 1 round max & 0.333 & 0.173 & 0.310 & {0.219} & {0.036} & 0.831 & {\textbf{0.479}} & {\textbf{0.197}} & 0.309 & {\textbf{0.485}} & {\textbf{0.121}} & \textbf{0.857} \\

\bottomrule
\end{tabular}
}
\caption{
Results for the \textbf{decentralized} and \textbf{centralized} Multi-LLM approaches. For the multi-LLM pipelines participating models are GPT-3.5 and GPT-4o mini.
The results use GPT-3.5 for the evaluator in the centralized approach, and summaries from GPT-3.5 are chosen in tie-breaking for both centralized and de-centralized approaches.
}

\label{tab:main-results}
\end{table*}

\subsection{Main Results} \label{sec:main-results}

Our multi-LLM framework outperforms single-LLM baselines by up to 3$\times$, as seen in Table~\ref{tab:main-results}. The fact that both precision- and recall-focused metrics improved means the multi-LLM approach is robust.
On average the centralized method improves the scores by 73\%, and the decentralized method outperforms baselines by 70\%. In our theoretical cost analysis (Section~\ref{sec:cent-analysis-and-discussion} and ~\ref{sec:decent-analysis-and-discussion}) we show that the input cost (in number of tokens) for the decentralized multiplies by the the number of agents participating in the evaluation, and with the more cost-effective centralized method our system is able to perform better than the single-LLM setup. This demonstrates the effectiveness of our proposed method under decentralized and decentralized frameworks.

We see that additional LLMs do not improve upon the 2-LLM setup (see Appendix~\ref{sec:exp-vary-num-LLMs}), and additional rounds of generation and evaluation do not further improve scores. This shows that even with just 2 LLMs and a single round of generation and evaluation we observe performance gains, meaning that the least costly version of the multi-LLM system is still able to deliver better summaries compared to single-LLM approaches.

In Table~\ref{tab:main-results} we use GPT-3.5 as the evaluator and tie-breaking choice in our multi-LLM. We also run the multi-LLM system with GPT-4o mini as the evaluator and the tie-breaker, and the results are shown in Table~\ref{tab:main-results-variance-gpt4oandgpt4ominiasevaluator}. Again, the multi-LLM framework outperformed single-LLM baselines, averaging 64\% improvement for the decentralized variant and 63\% for the centralized variant. In some individual scores, our framework improves upon single-LLM setups by up to 3$\times$. These improvements are competitive to those we obtain from the multi-LLM setup in Table~\ref{tab:main-results}, which means our proposed framework works well for different central models and different tie-breaking models.

We also perform additional experiments with other variables. More specifically, we assess the performance of the multi-LLM framework with alternative combinations of models, with three models contributing to the summarization and evaluation, and with models receiving fine-grained prompts instead of the same prompt. In all of these experiments, we obtain competitive results compared to the first decentralized and centralized setup, and the scores are higher than single-LLM baselines, showing that our proposed framework performs consistently under different setups.

\begin{table*}[htp]

\centering
\resizebox{0.9\textwidth}{!}{
\begin{tabular}{lcl ccHccH HHHHHH}
\toprule
 & \textbf{Max Rounds}
 & \textbf{Multi-LLM Model Combination} & ROUGE-1 $\uparrow$ & ROUGE-L $\uparrow$ & METEOR & BLEU-1 $\uparrow$ & BLEU-4 $\uparrow$ & BERT Score $\uparrow$ & ROUGE-1 $\uparrow$ & ROUGE-L $\uparrow$ & METEOR $\uparrow$ & BLEU-1 $\uparrow$ & BLEU-4 $\uparrow$ & BERT Score $\uparrow$\\
\midrule
\multirow{6}{*}{\bf Decentralized}
& \multirow{3}{*}{\bf 3 Rounds}
& \textcolor{googleblue}{\bf GPT-3.5} \& GPT-4o mini & \textbf{0.313} & \textbf{0.163} & \bf{0.300} & \textbf{0.200} & \bf{0.029} & 0.000 & 0.000 & 0.000 & 0.000 & 0.000 & 0.000 & 0.000 \\
&& GPT-4o \& \textcolor{googlegreen}{\bf GPT-3.5} & \textbf{0.313} & 0.159 & 0.294 & 0.197 & 0.025 & 0.827 & 0.000 & 0.000 & 0.000 & 0.000 & 0.000 & 0.000\\
&& GPT-4o \& \textcolor{googlegreen}{\bf GPT-4o mini} & 0.302 & 0.152 & 0.294 & 0.185 & 0.022 & 0.000 & 0.000 & 0.000 & 0.000 & 0.000 & 0.000 & 0.000\\
\cmidrule{2-12}
& \multirow{3}{*}{\bf 1 Rounds}
& \textcolor{googleblue}{\bf GPT-3.5} \& GPT-4o mini & \underline{\textbf{0.339}} & \underline{\textbf{0.180}} & \textbf{0.314} & \textbf{0.224} & \underline{\textbf{0.043}} & 0.000 & 0.000 & 0.000 & 0.000 & 0.000 & 0.000 & 0.000\\
&& GPT-4o \& \textcolor{googlegreen}{\bf GPT-3.5} & 0.328 & 0.170 & 0.304 & 0.212 & 0.033 & 0.000 & 0.000 & 0.000 & 0.000 & 0.000 & 0.000 & 0.000\\
&& GPT-4o \& \textcolor{googlegreen}{\bf GPT-4o mini} & 0.305 & 0.153 & 0.293 & 0.189 & 0.023 & 0.000 & 0.000 & 0.000 & 0.000 & 0.000 & 0.000 & 0.000\\

\midrule
\multirow{6}{*}{\bf Centralized}
& \multirow{3}{*}{\bf 3 Rounds}
& \textcolor{googlegreen}{\bf GPT-3.5} \& GPT-4o mini & \bf{0.329} & \bf{0.168} & \bf{0.302} & \bf{0.217} & \bf{0.031} & 0.000 & 0.000 & 0.000 & 0.000 & 0.000 & 0.000 & 0.000\\
&& GPT-4o \& \textcolor{googleblue}{\bf GPT-3.5} & 0.325 & 0.166 & 0.296 & 0.214 & 0.029 & 0.000 & 0.000 & 0.000 & 0.000 & 0.000 & 0.000 & 0.000\\
&& GPT-4o \& \textcolor{googleblue}{\bf GPT-4o mini} & 0.304 & 0.153 & 0.294 & 0.188 & 0.022 & 0.000 & 0.000 & 0.000 & 0.000 & 0.000 & 0.000 & 0.000\\
\cmidrule{2-12}
& \multirow{3}{*}{\bf 1 Rounds}
& \textcolor{googlegreen}{\bf GPT-3.5} \& GPT-4o mini & 0.333 & \bf{0.173} & \bf{0.310} & 0.219 & \bf{0.036} & 0.000 & 0.000 & 0.000 & 0.000 & 0.000 & 0.000 & 0.000\\
&& GPT-4o \& \textcolor{googleblue}{\bf GPT-3.5} & \underline{\textbf{0.339}} & 0.177 & 0.306 & \underline{\textbf{0.228}} & 0.039 & 0.000 & 0.000 & 0.000 & 0.000 & 0.000 & 0.000 & 0.000\\
&& GPT-4o \& \textcolor{googleblue}{\bf GPT-4o mini} & 0.306 & 0.155 & 0.294 & 0.190 & 0.022 & 0.000 & 0.000 & 0.000 & 0.000 & 0.000 & 0.000 & 0.000\\

\bottomrule

\end{tabular}
}
\caption{Varying the combination of models in our Multi-LLM approaches. 
Note rounds is the max number of rounds allowed and all results are for ArXiv. 
Bolded numbers are best scores for each round-model combination. Underlined numbers are overall best scores for each metric in this table. 
Furthermore, the central LLM is highlighted in \textcolor{googleblue}{blue} and for the decentralized multi-LLM approaches, we highlight the LLM used for tie-breaking in \textcolor{googlegreen}{green}.
}
\label{tab:varying-model-combinations}

\end{table*}

\subsection{Ablation Studies} \label{sec:main-text-ablation}

\textbf{Varying Model Combinations}
In Table~\ref{tab:main-results} we use GPT-3.5 and GPT-4o mini as the participating models in the multi-LLM framework. We further experiment with alternative combinations of models in the framework. As shown in Table~\ref{tab:varying-model-combinations} we again observe improvements across the board compared to the single-LLM baselines in Table~\ref{tab:main-results}, regardless of default model and number of rounds and type of interaction (decentralized vs. centralized). The improvements are competitive with those seen in the GPT-3.5 and GPT-4o mini combination. Further results are provided in Appendix~\ref{sec:exp-varying-model-comb}.

\textbf{Varying the Number of LLMs}
In this experiment we use 3 LLMs in the setup instead of 2. We observe a 54\% improvement for the decentralized method and 59\% for the centralized method on average over single-LLM summaries, and for individual scores we see improvements of up to 2.9$\times$. More detailed results are presented in Table~\ref{tab:vary-num-LLMs} and in Appendix~\ref{sec:exp-vary-num-LLMs}.

\textbf{Specialized Prompting}
In all previous experiments we have kept the generation prompt identical for all LLMs. With multi-LLM approaches, this need not be the case. In this experiment we choose different prompts for different models when generating summaries, aiming to have unique knowledge bases of different models complement each other. As seen in Table~\ref{tab:tableformultisummarymultillmfinal}, the centralized method results in a 66\% performance increase over single-LLM baselines in Table~\ref{tab:main-results}, and the decentralized method has a 58\% increase over the single-LLM baselines. For experimental details and further analysis see Section~\ref{sec:exp-single-LLM} in the Appendix

\textbf{Short vs. Long-text Multi-LLM Summarization}
In this experiment, we use only the introduction section as the basis for summarization in the ArXiv dataset. Since the introduction typically shorter than the context window of LLMs, we refer to these as "short-text" summarization, in contrast to the "long-text" summarization we explore previously. The results in Table~\ref{tab:main-results-shortsummarization} shows that the centralized approach provides the most performance gains over single-LLM baselines -- up to 2.4$\times$ on average, and the decentralized method sees a 2.3$\times$ increase. Further details can be found in Appendix~\ref{sec:short-long}.

\begin{table*}[htp]

\centering
\resizebox{0.9\textwidth}{!}{

\begin{tabular}{llrrrr}
\toprule
 &   & \textbf{Input Tokens} & \textbf{Output Tokens} & \textbf{Average Tokens} & \textbf{Total Tokens}\\
\midrule
\multirow{2}{*}{Decentralized} & Multi-LLM 3 round max & 383.73M & 25.63M & 14.62M & 409.37M\\
 & Multi-LLM 1 round max & 129.36M & 11.89M & 11.77M & 141.25M\\
\midrule
\multirow{2}{*}{Centralized} & Multi-LLM 3 round max & 216.65M & 19.55M & 14.76M & 236.2M\\
 & Multi-LLM 1 round max & 77.69M & 6.77M & 10.56M & 84.46M\\
\bottomrule
\end{tabular}

}

\caption{Cost Analysis of our Multi-LLM Decentralized and Centralized Summarization Methods. Note $M$= millions of tokens.}
\label{tab:cost-analysis}
\end{table*}
\subsection{Cost Analysis} \label{sec:exp-cost-analysis}

Table~\ref{tab:cost-analysis} presents the cost analysis for both decentralized and centralized methods based on the results in Table~\ref{tab:main-results}, highlighting key trends in input and output tokens across various stages of the summarization process. We observe that for evaluation stages the input and output token counts for the decentralized method are twice those for the centralized method, which reflect the number of LLMs in the setup.

\subsection{Human evaluation}
In addition to the ablation studies, we perform human evaluation of summaries similar to the last step of the multi-LLM framework, i.e. the evaluation phase (Section~\ref{sec:cent-convo-eval}). The human raters are prompted with two sets of summaries from the generation phase (Section~\ref{sec:cent-single-round-gen}), and are instructed to evaluate these two sets of summaries for Coherence, Conciseness, and Fluency on 5-point Likert scales \cite{conroy-dang-2008-mind} with rating guidelines for each possible score (Figure~\ref{fig:human-eval-prompt}). The summaries are randomized and anonymized to reduce bias attributable to knowledge of authorship. We drop the Relevance criteria since no original text is provided to the human raters due to length. 

We obtain 420 ratings from 7 raters, and find that humans generally prefer summaries produced by GPT-4o mini, which aligns with preferences by our multi-LLM framework. Human preferences also align with machine preferences for all three evaluation criteria to some degree. Conciseness has the highest agreement with multi-LLM evaluations ($\kappa=0.6$). More details for human evaluations can be found in Appendix~\ref{sec:human-evals}.

\section{Conclusion} \label{sec:conc}
This paper presented a multi-LLM framework for text summarization, and proposed two strategies, decentralized and centralized multi-LLM summarization.
We demonstrated that the proposed multi-LLM summarization techniques lead to better generated summaries.
Our results indicate that multi-LLM approaches are useful for improving text summarization.
Future work should continue to investigate multi-LLM approaches for summarization.

\section{Limitations} \label{sec:limit}
This work demonstrated the effectiveness of both our centralized and decentralized multi-LLM summarization approaches.
Future work should further investigate various aspects, including more diverse LLMs, and explore other topologies beyond the two extremes we proposed.
Furthermore, while we investigated a variety of datasets, future work can explore other domains.
We believe there are many approaches that lie between the two extreme multi-LLM strategies we investigated empirically in this work.
Finally, we did not optimize the prompts, as such we believe there is huge opportunity to achieve significantly better results by engineering better prompts to consider other important aspects of summarization.
We leave these and other important directions for future work.

\nocite{*}
\bibliography{main}
\bibliographystyle{acl_natbib}

\appendix

\section{Detailed Experimental Setup} \label{sec:exp-setup-detailed}

\noindent \textbf{Datasets:} We use the test sets of ArXiv \cite{cohan2018discourseawareattentionmodelabstractive} (first 20\%, or 1,288 documents) and GovReport \cite{huang-etal-2021-efficient} (all, or 973 documents) as document input for our summarization methods. They cover a range of genres, providing diverse texts for evaluation. In ArXiv, the main article excluding the abstract is the target for summarization, and the abstract is used as the ground truth reference summary; for GovReport, the text is the main report and the ground truth is the human-written summary. ArXiv articles range from 241 to 44,489 space-delimited words long, with an average of 5,950 words; their summaries range from 46 to 290 words, averaging 164 words. GovReport main texts range from 396 to 31,371 words, averaging 7,379 words; their summaries range from 67 to 1,363 words, averaging 571 words.

\noindent \textbf{Evaluation Metrics:} We assess the quality of LLM-generated summaries using ROUGE-1, ROUGE-L, BLEU-1, and BLEU-4 metrics. ROUGE scores emphasizes recall while BLEU scores emphasize precision.

\noindent \textbf{Baselines:} 
For comparison with our multi-LLM approach, unless otherwise mentioned, we leverage GPT-3.5, GPT-4o, GPT-4o mini, and LLaMA3-8B as baselines.
For these models, we perform the same chunking across all models, and the summarization prompt is identical to that in the first round of the multi-LLM summarization process (Figure~\ref{fig:initial-prompt-for-generating-summary}).
Unless otherwise mentioned, all models use 4K-char chunk-size, and the final summary for each document is concatenation of the generated summaries for each chunk in that document.

Finally, unless otherwise mentioned, we set $W=160$ for all models.

\section{Theoretical Analysis \& Discussion}
\subsection{Centralized Apporach} \label{sec:cent-analysis-and-discussion}

\paragraph{Cost and Complexity per Round} 
Let $I$ denote the number of input tokens in the original text and $O_{\max}$ represent an upper bound on the output tokens (i.e., maximum summary length). We consider $k$ distinct LLMs and a maximum of $t_{\max}$ conversational rounds. In each round $i$, we prompt all $k$ LLMs with approximately $I + \delta_i$ input tokens, where $\delta_i$ denotes additional tokens introduced in that round (e.g., references to previously generated summaries). Each LLM then produces up to $O_{\max}$ output tokens. Since input and output tokens often incur different costs, we consider them separately. For the generation phase, the input token cost per round is on the order of $\mathcal{O}(k \cdot (I + \delta_i))$, and the output token cost is on the order of $\mathcal{O}(k \cdot O_{\max})$. For evaluation, the central LLM processes $k$ candidate summaries and $I_{ec}$ instructions, resulting in an input token cost of about $\mathcal{O}(k \cdot O_{\max} + I_{ec})$. By directing the central LLM to output only an anonymous identifier for the chosen summary, we reduce output token length in evaluation, thereby minimizing the chance of hallucination and enabling more straightforward cost accounting.

\paragraph{Multi-Round Overhead} 
Over $t_{\max}$ rounds, the total input token usage for generation is $\mathcal{O}(t_{\max} \cdot k \cdot (I + O_{\max}))$, and the evaluation input token usage is $\mathcal{O}(t_{\max} \cdot (k \cdot O_{\max} + I_{ec}))$. Although this complexity may appear large, $t_{\max}$ is typically small (e.g., $2$ or $3$), and $O_{\max}$ is usually constrained (e.g., a brief 160-word summary). Moreover, careful prompt engineering can curtail $\delta_i$ growth, ensuring that the number of tokens per round remains bounded.

\paragraph{Convergence and Quality Gains} 
The iterative generation-evaluation mechanism aims to converge within a small number of rounds. With each iteration, models refine their outputs guided by previous results, potentially improving summary quality. This iterative refinement, while incurring additional steps, offers a practical trade-off between computation and quality, as the improved summaries can justify the limited number of extra rounds.

\subsection{Decentralized Approach} \label{sec:decent-analysis-and-discussion}

\paragraph{Multi-Round Complexity} 
Let $I$ denote the number of input tokens in the original text and $O_{\max}$ represent an upper bound on the output tokens (i.e., maximum summary length). We consider $k$ distinct LLMs and a maximum of $t_{\max}$ conversational rounds.

Over $t_{\max}$ rounds, the worst-case token cost from generation is:
\[
\mathcal{O}(t_{\max} \cdot k \cdot (I + O_{\max})).
\]
The evaluation cost scales to:
\[
\mathcal{O}(t_{\max} \cdot (k \cdot I_{e} + k^2 \cdot O_{\max})).
\]
Combined, we have a total complexity per round of approximately:
\[
\mathcal{O}(k \cdot I + k \cdot O_{\max} + k \cdot I_{e} + k^2 \cdot O_{\max}).
\]
Thus, for $t_{\max}$ rounds, the overall complexity becomes:
\[
\mathcal{O}(t_{\max} \cdot (k \cdot I + k \cdot I_{e} + k \cdot O_{\max} + k^2 \cdot O_{\max})).
\]
Since $k^2 \cdot O_{\max}$ may dominate for large $k$, this term can become the bottleneck. However, in practical scenarios, $k$ (the number of LLMs) is often small (e.g., $2$--$5$), making the decentralized evaluation overhead manageable.

\paragraph{Trade-Offs and Practical Considerations}
The decentralized evaluation approach increases computational overhead compared to the centralized model, as it requires every model to evaluate all candidate summaries. However, this additional cost is justified by the potential gains in robustness and reliability of the final output, but also by the flexibility to rely on multiple, potentially weaker models rather than a single, highly capable central evaluator. By employing a form of consensus voting, the system can arrive at a more stable decision even when no single model is individually strong.

While the added complexity of multi-round conversation can be non-trivial, it may lead to improved summary quality, especially when dealing with contentious or ambiguous source texts. Multiple rounds allow the system to refine the summaries and converge on a stable solution. If consensus emerges quickly, the number of rounds $t_{\max}$ can be effectively reduced, thereby decreasing the total computational cost. Conversely, if no consensus is reached, the algorithm ultimately defaults to a tie-break mechanism after $t_{\max}$ rounds, ensuring bounded time and cost. As with the centralized approach, prompt engineering and careful parameter selection (e.g., choosing $O_{\max}$, $t_{\max}$, and the number of participating models $k$) we can mitigate undue complexity.

\section{Ablation Study}
\subsection{Varying Evaluation LLM} \label{sec:exp-varying-centralized-llm}
In this section, we compare the scores of the centralized and decentralized approaches when the evaluator model and the tie-breaker models are GPT-4o or GPT-4o mini (instead of GPT-3.5 as in Table~\ref{tab:main-results}). These results are presented in Table~\ref{tab:main-results-variance-gpt4oandgpt4ominiasevaluator}. The sections where GPT-3.5 is the evaluator are reproduced from Table~\ref{tab:main-results}.

In these experiments, the summary-generating models remain the same as those in Table~\ref{tab:main-results}. In rows (in Table~\ref{tab:main-results-variance-gpt4oandgpt4ominiasevaluator}) where GPT-4o is listed as the evaluator, however, the decentralized method would have required GPT-4o to be the default choice for a tie-breaking summary as well when the model has not generated summaries. To remain maximally consistent with previous methodology, we modify the process here so that GPT-4o receives the final-round summaries from the decentralized method where GPT-3.5 is the tie-breaking choice and evaluator and performs a \textbf{centralized} evaluation on top of the decentralized results. The reason the GPT-3.5-default results are chosen as the basis instead of GPT-4o mini is because as an evaluator and default choice GPT-3.5 produced better final summaries compared to GPT-4o mini for both centralized and decentralized methods.

The multi-LLM framework outperformed single-LLM baselines, averaging 64\% improvement for the decentralized variant and 63\% for the centralized variant. In some individual scores, our framework improves upon single-LLM setups by up to 3$\times$. GPT-3.5 emerged as the best-scoring evaluator and the best-scoring tie-breaker choice: for the centralized method, GPT-3.5 as an evaluator and tie-breaking choice outperforms other evaluators and tie-breakers, and for the decentralized method, GPT-3.5 turned out to be the best tie-breaking choice. Furthermore, GPT-3.5 as a centralized evaluator and tie-breaking choice separately outperform \textbf{both the decentralized and centralized} methods using other models as the evaluator and tie-breaking choice. As with results in Table~\ref{tab:main-results}, additional rounds of evaluation and regeneration do not improve summary scores.

\subsection{Varying Model Combinations} \label{sec:exp-varying-model-comb}
In Table~\ref{tab:main-results} we present the results with GPT-3.5 and GPT-4o mini as the models in the combination; we now investigate the performance of our approaches for alternative combinations of LLMs (in Table~\ref{tab:varying-model-combinations}). We use the following combinations for the 2-LLM framework: GPT-3.5 and GPT-4o mini, with GPT-3.5 as the evaluator and default, GPT-4o and GPT-3.5, again with GPT-3.5 as evaluator and default, and finally GPT-4o and GPT-4o mini, with GPT-4o mini as the evaluator and default.

These alternative combinations all outperform single-LLM baselines. We see a 54\% improvement in the decentralized variant and a 59\% for the centralized variant. Combinations with GPT-3.5 as a member and the evaluator/default choice offer larger improvements compared to those without GPT-3.5. Since we have used GPT-4o mini as the evaluator and tie-breaker where GPT-3.5 is absent, a possible reason the improvements for these pairings are less than those where GPT-3.5 is present is that GPT-3.5 is a larger model than GPT-4o mini.

\subsection{Varying the Number of LLMs} \label{sec:exp-vary-num-LLMs}

In this experiment, we increase the number of LLMs in our multi-LLM system to ascertain the effects on summary quality, and present the results in Table~\ref{tab:vary-num-LLMs}. Here we use GPT-3.5, GPT-4o mini, and GPT-4o in the multi-LLM system. We see that while the 3-LLM system still outperform the single-LLM baseline, increasing the number of LLMs from 2 to 3 does not improve performance upon the 2-LLM system, contrary to the trend observed in the previous sections where 2-LLM system outperform single-LLM baselines.

We offer two possible explanations for this finding. First, adding an additional LLM increases the complexity of the pipeline, which may lead to propagation of noise or redundancy in intermediate summaries. This added complexity could dilute the strengths of individual LLMs and reduce overall coherence and relevance in the final output. Second, the integration of a third LLM introduces a greater risk of inconsistencies in summarization styles, which may negatively affect evaluation metrics like ROUGE that rely on lexical overlap.

\begin{table*}[htp]

\resizebox{1.0\textwidth}{!}{

\begin{tabular}{lll ccHcc ccHcc}
\toprule
 &  & & \multicolumn{5}{c}{\bf{ArXiv}} & \multicolumn{5}{c}{\bf{GovReport}} \\
\cmidrule(lr){4-8}
\cmidrule(l){9-13}
 &  & 
 & ROUGE-1 $\uparrow$ & ROUGE-L $\uparrow$ & METEOR & BLEU-1 $\uparrow$ & BLEU-4 $\uparrow$ 
 & ROUGE-1 $\uparrow$ & ROUGE-L $\uparrow$ & METEOR $\uparrow$ & BLEU-1 $\uparrow$ & BLEU-4 $\uparrow$ \\

\midrule
\multirow{4}{*}{\bf GPT-4o mini Evaluator}
&\multirow{2}{*}{\bf Decentralized}
& Multi-LLM 3 round max & 0.317 & 0.160 & 0.296 & 0.206 & 0.026 & \textbf{0.445} & \textbf{0.178} & 0.293 & \textbf{0.452} & \textbf{0.094} \\
& & Multi-LLM 1 round max & \textbf{0.326} & \textbf{0.163} & 0.295 & \textbf{0.221} & \textbf{0.027} & 0.438 & 0.175 & 0.288 & 0.446 & 0.089  \\
\cmidrule{2-13}
&\multirow{2}{*}{\bf Centralized}
& Multi-LLM 3 round max & 0.315 & 0.158 & 0.300 & 0.201 & 0.027 & \textbf{0.441} & \textbf{0.176} & 0.292 & \textbf{0.447} & \textbf{0.092} \\
& & Multi-LLM 1 round max & \textbf{0.330} & \textbf{0.165} & 0.297 & \textbf{0.222} & \textbf{0.028} & 0.439 & 0.175 & 0.290 & 0.446 & 0.090  \\

\midrule
\multirow{4}{*}{\bf GPT-3.5 Evaluator} &
\multirow{2}{*}{\bf Decentralized} 
& Multi-LLM 3 round max & 0.313 & 0.163 & 0.301 & 0.200 & 0.029 & 0.447 & 0.180 & 0.292 & 0.458 & 0.098 \\
& & Multi-LLM 1 round max & \textbf{\underline{0.339}} & \textbf{\underline{0.180}} & \textbf{0.315} & \textbf{\underline{0.224}} & \textbf{\underline{0.043}} & \textbf{0.468} & \textbf{0.190} & 0.305 & \textbf{0.477} & \textbf{0.112} \\
\cmidrule{2-13}
&\multirow{2}{*}{\bf Centralized} 
& Multi-LLM 3 round max & 0.329 & 0.168 & 0.302 & 0.217 & 0.031 & 0.468 & 0.189 & 0.307 & 0.470 & 0.109 \\
& & Multi-LLM 1 round max &  \textbf{0.333} &  \textbf{0.173} & 0.310 & \textbf{0.219} & \textbf{0.036} & \textbf{\underline{0.479}} & \textbf{\underline{0.197}} & 0.309 & \textbf{\underline{0.485}} & \textbf{\underline{0.121}} \\

\midrule

\multirow{4}{*}{\bf GPT-4o Evaluator} &
\multirow{2}{*}{\bf Decentralized}
& Multi-LLM 3 round max & \textbf{0.326} & \textbf{0.166} & 0.302 & \textbf{0.214} &0.030 & 0.446 & 0.179 & 0.293 &0.456 &0.098 \\
& & Multi-LLM 1 round max & 0.325 & 0.165 & 0.304 & 0.211 & 0.030 & \textbf{0.456} & \textbf{0.183} & 0.300 & \textbf{0.461} & \textbf{0.100} \\
\cmidrule{2-13}
&\multirow{2}{*}{\bf Centralized}
& Multi-LLM 3 round max & 0.318 & 0.162 & 0.300 & 0.206 & 0.027  & 0.449 & 0.181 & 0.298 & 0.452 & 0.096 \\
& & Multi-LLM 1 round max & \textbf{0.327} & \textbf{0.167} & 0.303 & \textbf{0.215} & \textbf{0.031}  & \textbf{0.461} & \textbf{0.186} & 0.303 & \textbf{0.467} & \textbf{0.105} \\
\bottomrule

\end{tabular}
}
\caption{
Results for different evaluating and tie-breaking models for Multi-LLM approaches. The choice of the tie-breaker models is the same as the choice of evaluator model. We bold the best results for each combination of the experimental variables, and we underline the best results overall. For ease of comparison, we reproduce the best-performing 2-LLM results obtained in Table~\ref{tab:main-results}}

\label{tab:main-results-variance-gpt4oandgpt4ominiasevaluator}

\end{table*}

\begin{table*}[htp]

\resizebox{1.0\textwidth}{!}{

\begin{tabular}{lll ccHccH ccHccH}

\toprule
 &  & & \multicolumn{6}{c}{\bf{ArXiv}} & \multicolumn{6}{c}{\bf{GovReport}} \\
\cmidrule(lr){4-8}
\cmidrule(lr){9-15}
 &  & 
 & ROUGE-1 $\uparrow$ & ROUGE-L $\uparrow$ & METEOR & BLEU-1 $\uparrow$ & BLEU-4 $\uparrow$ & BERT Score $\uparrow$ & ROUGE-1 $\uparrow$ & ROUGE-L $\uparrow$ & METEOR $\uparrow$ & BLEU-1 $\uparrow$ & BLEU-4 $\uparrow$ & BERT Score $\uparrow$\\

\midrule

\multirow[c]{4}{*}{\shortstack[l]{\bf 2-LLMs\\GPT-3.5 Evaluator}}
& \multirow[c]{2}{*}{\bf Decentralized} & 3 rounds & 0.313 & 0.163 & 0.301 & 0.200 & 0.029 & 0.829 & 0.447 & 0.180 & 0.292 & 0.458 & 0.098 & 0.855 \\
 && 1 rounds & \textbf{\underline{0.339}} & \textbf{\underline{0.180}} & \textbf{\underline{0.315}} & \textbf{\underline{0.224}} & \textbf{\underline{0.043}} & \textbf{\underline{0.832}} & \textbf{0.468} & \textbf{0.190} & \textbf{0.305} & \textbf{0.477} & \textbf{0.112} & \textbf{\underline{0.857}} \\
 
\cmidrule{2-15}

& \multirow[c]{2}{*}{\bf Centralized} & 3 rounds & 0.329 & 0.168 & 0.302 & 0.217 & 0.031 & 0.830 & 0.468 & 0.189 & 0.307 & 0.470 & 0.109 & \textbf{\underline{0.857}} \\
 && 1 rounds & \textbf{0.333} & \textbf{0.173} & \textbf{0.310} & \textbf{0.219} & \textbf{0.036} & \textbf{0.831} & \textbf{\underline{0.479}} & \textbf{\underline{0.197}} & \textbf{0.309} & \textbf{\underline{0.485}} & \textbf{\underline{0.121}} & \textbf{\underline{0.857}} \\
 
\midrule

\multirow[c]{4}{*}{\shortstack[l]{\bf 3-LLMs\\GPT-4o mini Evaluator}}
 & \multirow[c]{2}{*}{\bf Decentralized} 
  & 3 rounds & \textbf{0.301} & \textbf{0.154} & \textbf{0.297} & 0.184 & \textbf{0.024} & \textbf{0.827} & \textbf{0.445} & 0.178 & \textbf{0.297} & \textbf{0.449} & \textbf{0.095} & 0.854 \\
 & & 1 rounds & 0.299 & 0.152 & 0.295 & 0.184 & 0.023 & 0.826 & 0.442 & 0.178 & 0.293 & 0.447 & 0.094 & 0.854 \\
 
\cmidrule{2-15}

 & \multirow[c]{2}{*}{\bf Centralized} 
  & 3 rounds & 0.300 & \textbf{0.153} & \textbf{0.295} & 0.185 & 0.023 & 0.826 & \textbf{0.443} & 0.178 & 0.293 & 0.447 & \textbf{0.094} & 0.854 \\
 & & 1 rounds & 0.300 & 0.152 & 0.295 & \textbf{0.186} & 0.023 & 0.826 & 0.442 & 0.178 & 0.291 & \textbf{0.449} & 0.093 & 0.854 \\

\midrule

\multirow[c]{4}{*}{\shortstack[l]{\bf 3-LLMs\\GPT-3.5 Evaluator}}
 & \multirow[c]{2}{*}{\bf Decentralized} 
 & 3 rounds & 0.300 & 0.154 & 0.297 & 0.184 & 0.024 & 0.827 & 0.446 & 0.179 & 0.298 & 0.443 & 0.094 & 0.855 \\
 & & 1 rounds & \textbf{0.309} & \textbf{0.159} & \textbf{0.303} & \textbf{0.193} & \textbf{0.027} & \textbf{0.830} & \textbf{0.451} & \textbf{0.182} & 0.287 & \textbf{0.459} & \textbf{0.099} & \textbf{0.856} \\
 
\cmidrule{2-15}

 & \multirow[c]{2}{*}{\bf Centralized} 
  & 3 rounds & 0.294 & 0.151 & 0.295 & 0.177 & 0.023 & 0.827 & 0.451 & 0.181 & \textbf{0.302} & 0.440 & 0.095 & 0.855 \\
 & & 1 rounds & \textbf{0.329} & \textbf{0.172} & \textbf{0.307} & \textbf{0.214} & \textbf{0.036} & \textbf{0.830} & \textbf{0.460} & \textbf{0.189} & 0.291 & \textbf{0.451} & \textbf{0.104} & 0.855 \\

\bottomrule

\end{tabular}
}

\caption{Multi-LLM framework with three models. We bold the best results for each combination of the experimental variables, and we underline the best results overall. For ease of comparison, we reproduce the best-performing 2-LLM results obtained in Table~\ref{tab:main-results}}
\label{tab:vary-num-LLMs}

\end{table*}

\subsection{Specialized Prompting} \label{sec:exp-single-LLM}

We now investigate using a single LLM to generate 
multiple different summaries of the text, and then using our framework to obtain the best summary. 
We explore the efficacy of varying prompt formulations and model parameters in regards to our framework. This experiment is grounded in the intuition that long documents contain very diverse sections within their content which may benefit from different summarization strategies. For example, different chunks of a long document may cover distinct topics, serve various purposes, have diverse writing styles, and/or contain differing density. Given this diversity, a simple uniform summarization prompt is less likely to actually capture the required essential information from each chunk. With this, we propose a form of specialized prompting as a way to leverage the distinctive capabilities and specializations of each model for specific chunks specifically in regards to our framework. We hypothesize that the use of specialized prompting can help further leverage LLM capabilities within our suggested multi-LLM framework to produce higher quality summaries which are more suitable for subsequent evaluation by multiple LLMs.

We begin by generating four initial summaries using two sets of specialized prompts designed for GPT-3.5 and GPT-4o mini, ensuring that each model receives two distinct prompts. One prompt focuses on enhancing the coherence of the resulting summary (see Figure~\ref{fig:initial-prompt-for-generating-summary1-specialized}), while the second prompt aims to maximize precision in conveying the key facts (see Figure~\ref{fig:initial-prompt-for-generating-summary2-specialized}). After producing these four baseline summaries, we feed them into our multi-LLM framework, which incorporates two agents — GPT-3.5 and GPT-4o mini—working collaboratively. GPT-3.5 and GPT-4o mini are used for the initial generation of summaries, and GPT-3.5 also serves as the evaluator. The framework and methodology following the generation of the four baseline summaries, as well as their inclusion as input, mirror the procedures used to obtain decentralized and centralized results on ArXiv and GovReport in Table~\ref{tab:main-results}, with GPT-3.5 functioning as the evaluator. Results for this experiment are provided in Table~\ref{tab:tableformultisummarymultillmfinal}.

This experiment demonstrates that employing specialized prompting strategies within both decentralized and centralized multi-LLM frameworks significantly enhances the quality of generated summaries. These results show the importance of prompt engineering and strategic framework design in multi-LLM summarization tasks and we leave this for future work.

\begin{table*}[htp]

\resizebox{1.0\textwidth}{!}{
\begin{tabular}{lllrrHrrHrrHrrH}
\toprule
 &  & & \multicolumn{6}{c}{\bf{ArXiv}} & \multicolumn{6}{c}{\bf{GovReport}} \\
 &  & & ROUGE-1 ↑ & ROUGE-L ↑ & METEOR ↑ & BLEU-1 ↑ & BLEU-4 ↑ & BERT Score ↑
 & ROUGE-1 ↑ & ROUGE-L ↑ & METEOR ↑ & BLEU-1 ↑ & BLEU-4 ↑ & BERT Score ↑\\
\midrule

\multirow[c]{4}{*}{\shortstack[l]{\bf Baseline Prompts}}
& \multirow[c]{2}{*}{\bf Decentralized} & 3 round max & 0.313 & 0.163 & 0.301 & 0.200 & 0.029 & 0.829 & 0.447 & 0.180 & 0.292 & 0.458 & 0.098 & 0.855 \\
 &&  1 round max & \textbf{0.339} & \textbf{0.180} & \textbf{0.315} & \textbf{0.224} & \textbf{0.043} & \textbf{0.832} & \textbf{0.468} & \textbf{0.190} & \textbf{0.305} & \textbf{0.477} & \textbf{0.112} & \textbf{0.857} \\
 
\cmidrule{2-15}

& \multirow[c]{2}{*}{\bf Centralized} & 3 round max & 0.329 & 0.168 & 0.302 & 0.217 & 0.031 & 0.830 & 0.468 & 0.189 & 0.307 & 0.470 & 0.109 & \textbf{0.857} \\
 && 1 round max & \textbf{0.333} & \textbf{0.173} & 0.310 & \textbf{0.219} & \textbf{0.036} & 0.831 & \textbf{0.479} & \textbf{\underline{0.197}} & 0.309 & \textbf{0.485} & \textbf{\underline{0.121}} & \textbf{0.857} \\

\midrule

\multirow[c]{4}{*}{\shortstack[l]{\bf Specialized Prompts}}
 & \multirow[c]{2}{*}{\bf Decentralized} 
  &  3 round max & 0.300 & 0.155 & 0. & 0.201 & 0.025 & 0.827 & 0.464 & 0.174 & 0.300 & 0.441 & 0.093 & 0.854 \\
 & &  1 round max & \textbf{0.338} & \textbf{0.175} & 0.306 & \textbf{0.236} & \textbf{0.040} &\textbf{ 0.830} & \textbf{0.469} & \textbf{0.181} &\textbf{ 0.295} & \textbf{0.486} & \textbf{0.104} & 0.856 \\

\cmidrule{2-15}

 & \multirow[c]{2}{*}{\bf Centralized} 
  &  3 round max & 0.316 & 0.162 & 0.297 & 0.215 & 0.032 & 0.828 & 0.473 & 0.177 & 0.301 & 0.452 & 0.101 & 0.856 \\
 & &  1 round max & \textbf{\underline{0.355}} & \textbf{\underline{0.181}} & 0.310 & \textbf{\underline{0.251}} &\textbf{\underline{0.049}} & 0.831 & \textbf{\underline{0.482}}& \textbf{0.185} & 0.297 & \textbf{\underline{0.494}} & \textbf{0.115} & 0.856\\

\bottomrule

\end{tabular}
}
\caption{Results on the use of 2 specialized prompts on where the only change in the pipeline is that 4 total specialized baseline summaries are fed in initially instead of the 2 simple prompts fed in the methodology used to curate Table~\ref{tab:main-results}. Note that these results use GPT-3.5 for the evaluator in the centralized approach, and for breaking ties in the decentralized multi-LLM approaches. This is for a 15 sample size for both datasets. Refer to Figure~\ref{fig:initial-prompt-for-generating-summary1-specialized} and Figure~\ref{fig:initial-prompt-for-generating-summary2-specialized} for the prompts used for initial generation. We bold the best results for each combination of the experimental variables, and we underline the best results overall.}

\label{tab:tableformultisummarymultillmfinal}

\end{table*}

\begin{figure}[t!]
        \begin{formal}
        \small
        \textit{\tt 
        \\
        Generate a summary that enhances coherence of the text in around 160 words. Output the summary text only and nothing else.
        \begin{center}
        [text]
        \end{center}
        }
        \end{formal}
        \caption{Prompt 1 for generating the initial summary in the first round.}
        \label{fig:initial-prompt-for-generating-summary1-specialized}
        \end{figure}

\begin{figure}[t!]
        \begin{formal}
        \small
        \textit{\tt 
        \\
        Generate a summary that maximizes precision related to the key facts of the text in around 160 words. Output the summary text only and nothing else.
        \begin{center}
        [text]
        \end{center}
        }
        \end{formal}
        \caption{Prompt 2 for generating the initial summary in the first round.}
        \label{fig:initial-prompt-for-generating-summary2-specialized}
        \end{figure}

\subsection{Short-text vs. Long-text Summarization}\label{sec:short-long}

\begin{figure*}[ht]
    \centering
    \fbox{\includegraphics[width=\textwidth]{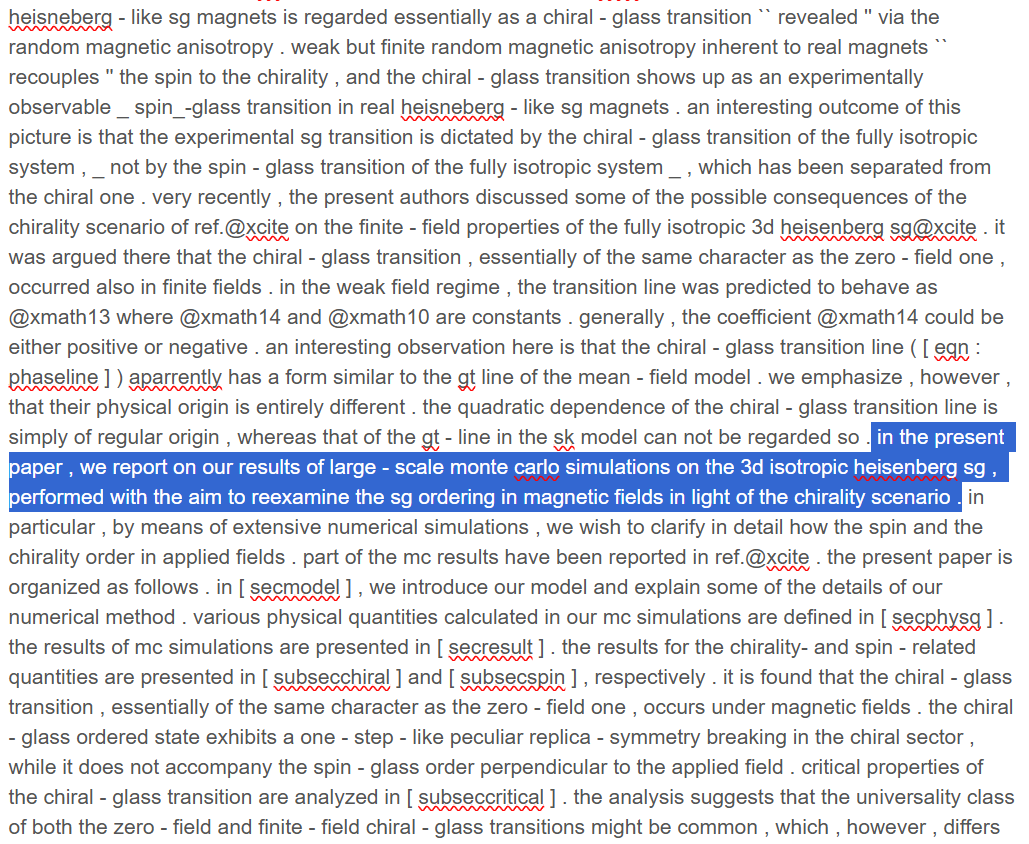}}
    \caption{Here we showcase an example of how we choose at which point an introduction ends. The total word count of this example article was 7,671 and the word count of the reference summary was 172. We highlight the inflection sentence which most serves as the transition from the actual background and theoretical setup of the paper to the actual methodologies which are then detailed in later text. From here we gather the word count of everything before the inflection sentence and classify it as our reference introduction text for experimentation, including the inflection sentence. In this case, the resulting introduction section had a total word count of 1203. }
    \label{fig:summarization_comparison}
\end{figure*}

In this section, we investigate the effectiveness of our approach for shorter text summarization.
For this experiment, we leverage the ArXiv dataset and only use the introduction of the paper as input for summarization and evaluate against the same ground-truth. The introduction subsections of papers are typically rich in content yet contain enough brevity to serve as quality standardized reference bases for our goal of long and short text experimentation.
With this experiment we present results that showcase the trade offs and performance differences of our methodologies on shorter text summarization compared to that of long document summarization. Generally, ArXiv papers contain detailed markers and section titles to distinguish introduction sections. However, using the Hugging Face dataset of ArXiv papers for our experimentation the format in which the article is represented is a string containing the "body" of the paper which contains little to no explicit markers for section identification. Thus, we present a simple heuristic to distinguish the introduction text from the rest of the article text. We manually went through 5 randomized example articles, with an assumption that the beginning of the article text starts with the introduction section, and found at which inflection point the introduction section concludes. After averaging the word count of the introduction sections and including an extension buffer to capture certain articles which may have slightly longer introduction sections we establish a benchmark for the using the first 1,500 words in ArXiv articles as our reference introduction section. We algorithmically consider a word as a break between the article string wherever there is whitespace. Refer to Figure~\ref{fig:summarization_comparison} for more detailed explanation. We ultimately curate 20\% of the examples from the test set using this strategy for performance testing on our metrics. Full results are provided in Table~\ref{tab:main-results-shortsummarization}.

\begin{table*}[htp]

\centering
\resizebox{\textwidth}{!}{%
\begin{tabular}{clcccHcc}
\toprule
 &  & \multicolumn{3}{c}{\bf ArXiv} \\
 &  & & {ROUGE-1 ↑} & {ROUGE-L ↑} & {METEOR} & {BLEU-1 ↑} & {BLEU-4 ↑} \\
\midrule
\midrule
\multirow{4}{*}{Long Text} & \multirow{2}{*}{\bf Decentralized} & Multi-LLM 3 round max & 0.329 & 0.168 & 0.302 & 0.217 & 0.031 \\
 & & Multi-LLM 1 round max & \textbf{0.333} & \textbf{0.173} & 0.310 & \textbf{0.219} & \textbf{0.036} \\
\cmidrule{2-8}
 & \multirow{2}{*}{\bf Centralized} & Multi-LLM 3 round max & 0.313 & 0.163 & 0.301 & 0.200 & 0.029 \\
 & & Multi-LLM 1 round max & \textbf{0.338 }& \textbf{0.180} & \textbf{0.315} &\textbf{ 0.224} & \textbf{0.043} \\
\midrule
\multirow{4}{*}{Short Text} & \multirow{2}{*}{\bf Decentralized} & Multi-LLM 3 round max & 0.360 & 0.188 & \bf0.247 & \bf\underline{0.328} & 0.038 \\
 & & Multi-LLM 1 round max & \textbf{0.369} & \textbf{0.198} & 0.242 & 0.309 & \textbf{0.044} \\
\cmidrule{2-8}
 & \multirow{2}{*}{\bf Centralized} & Multi-LLM 3 round max & 0.367 & 0.194 & 0.245 & \textbf{0.321} & 0.041 \\
 & & Multi-LLM 1 round max & \textbf{\underline{0.379}} & \textbf{\underline{0.206}} & 0.244 & 0.305 & \textbf{\underline{0.049}} \\
\bottomrule
\end{tabular}%
}
\caption{Results on short summarization tasks using the ArXiv dataset for the decentralized and centralized Multi-LLM approaches. Note that these results use
GPT-3.5 for the evaluator in the centralized approach, and for breaking ties in the decentralized multi-LLM
approaches.}

\label{tab:main-results-shortsummarization}

\end{table*}

We highlight several key aspects of our multi-LLM summarization methodology using both the centralized and decentralized approaches, showcasing distinct performance across both long and short text summarization tasks.As evident by our results, short articles consistently show better performance compared to long articles, showcasing the inherent complexities and nuances of longer texts that plague LLMs in terms of capturing and summarizing relevant content. 
The similar performance on metrics like ROUGE-1 and BLEU-4 in our centralized approach across different text lengths might indicate a consistency in how our methodology is able to capture the essential content and has the ability to reproduce the core narrative elements of the text regardless of length. Furthermore, we posit the difference in performance across long and short text for BLEU-1 is based primarily on the metrics itself as it measures the unigram overlap between the generated summary and the reference text. In the case of short texts, the decentralized approach and especially the 3 round performs best as each round and each model provides an opportunity to focus more accurately on and determine crucial unigrams that are significant within the context of a compact introduction section. This iterative refinement likely leads to a higher precision in capturing key terms and phrases, directly contributing to better BLEU-1 scores than in the case of the centralized approach which performs best as the context length is scaled up as shown in the results for long text.

\section{Human Evaluation}
\label{sec:human-evals}
\begin{figure*}[ht]

\centering
\fbox{\includegraphics[width=\textwidth]{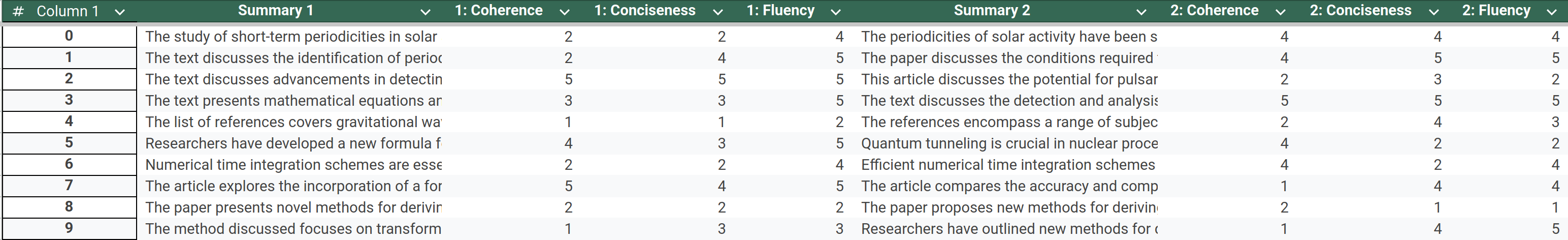}}

\begin{formal}
\small
Read the two summaries (column C and H) and grade them on the following aspects 
Give a grade from 1 - 5 for each category, 1 being the lowest score and 5 being the highest score, as described below. For summary 1, fill in columns E-G; for summary 2, fill in columns J-L.\\

\textbf{Coherence}\\
Evaluate the logical flow of sentences and organization of information within the summary.\\
1: Ideas do not logically follow from one to another, even if they are relevant.\\
2: There are some phrases to connect one idea to another.\\
3: Main ideas connect logically with each other. A few arguments/points are out of order.\\
4: Most ideas follow logically from one to the next. There may be minor points or words/phrases that seem out of place.\\
5: All ideas, even if irrelevant, are well connected and organized in a clear overall structure.\\

\textbf{Conciseness}\\
Penalize summaries that are overly verbose.\\
1: Ideas are repeated multiple times, or are described in excessive detail or are verbose, even if ideas are logically organized or are relevant.\\
2: Ideas/arguments/phrases are sometimes repeated. Most ideas are described in too much detail, or, the text is generally verbose with jargon.\\
3: Ideas are generally described in appropriate detail. Parts of texts may be unnecessarily verbose or use unnecessary jargon.\\
4: Most ideas are described in appropriate detail. There may be occasional verbosity.\\
5: All ideas are described with appropriately complex or simple sentences.\\

\textbf{Fluency}\\

Evaluate the grammatical correctness of the generated summary.\\
Check for any awkward phrasing or grammatical errors that might hinder comprehension.
1: Text is grammatically incorrect or very difficult to understand. There are incomplete sentences or incorrectly used punctuations.\\
2: The text is generally difficult to understand, but some sections convey meaningful ideas.\\
3: The text is generally grammatically correct. Some sentences may have incorrect grammar.\\
4: The text is grammatically correct and sentences have understandable structure. There are occasional incorrectly phrased sentences.\\
5: Summary is easy to follow and understand. No grammatical errors.
\end{formal}
\caption{Screenshot of the interface (with scores already filled in) and instructions given to raters for human evaluation. Instructions include the column indices (not shown in the screenshot) for easier reference. Each summary is rated according to the criteria listed in the instructions (i.e. Coherence, Conciseness, and Fluency). We provide guidelines for each criterion and each possible score for that criterion.}
\label{fig:human-eval-prompt}

\end{figure*}

In the human evaluation, we select the first 10 pairs of summaries generated before the final evaluation step by the decentralized, one-round maximum framework (the best-performing setup for ArXiv; see Table~\ref{tab:main-results}), and prompt human raters to rate each summary according to Coherence, Conciseness, and Fluency metrics, each represented by a 5-point Likert scale. The goal is to compare human rater preferences with preferences of the LLM performing the final evaluation and therefore producing the final result of the multi-LLM pipeline. Each summary pair consists of texts generated by GPT-3.5 and GPT-4o mini randomized in order of presentation and anonymized such that the raters do not know which model produced which summaries. We do not prompt raters with the corresponding original text as in the multi-LLM method due to the original text's length and technicality, and therefore we remove the Relevance criterion used in \citeauthor{conroy-dang-2008-mind} (\citeyear{conroy-dang-2008-mind}). For the remaining criteria we provide guidelines for each possible point value to improve reproducibility. Instructions provided to human raters and rating guidelines can be found in Figure~\ref{fig:human-eval-prompt}.

At the end of human evaluation, we collect 420 ratings from 7 raters (6 men, 1 woman; ages 23-31; 4 East Asian, 2 South Asian, 1 White).\footnote{Raters are authors of this paper and close associates who have consented to submitting evaluations, and no rater has prior knowledge of authors of the set of summaries being evaluated.}
To determine preferences for the human raters, we average the rating for each model and each summary in each evaluation criterion (Table~\ref{tab:human-eval-stats1}). Thus, for each summary and for each criterion we obtain two averaged scores, one for GPT-3.5 and GPT-4o mini. We then determine the human preference by choosing the model with the higher score, and if the scores are the same, we fallback on the default choice GPT-3.5, consistent with the fallback default in the evaluation step in our multi-LLM framework (Table~\ref{tab:human-eval-choices}). We note that in all three criteria our framework show some agreement with the human raters, as measured by Cohen's kappa. For conciseness, we observe an agreement of $\kappa = 0.6$.

\begin{table*}[htp]
\centering
\resizebox{\textwidth}{!}{%
\begin{tabular}{rcccccccc}
\hline
\textbf{Summary} & \multicolumn{2}{c}{\textbf{Coherence}} & \multicolumn{2}{c}{\textbf{Conciseness}} & \multicolumn{2}{c}{\textbf{Fluency}} & \multicolumn{2}{c}{\textbf{Averaged}} \\
Average scores\\ by raters for: & \textbf{GPT-3.5} & \textbf{GPT-4o mini} & \textbf{GPT-3.5} & \textbf{GPT-4o mini} & \textbf{GPT-3.5} & \textbf{GPT-4o mini} & \textbf{GPT-3.5} & \textbf{GPT-4o mini} \\
\hline
1 & \underline{3.57} & 3.57 & \underline{\textbf{3.85}} & 3.57 & \underline{\textbf{4.42}} & 4.28 & \underline{\textbf{3.95}} & 3.80\\
2 & \underline{\textbf{4.28}} & 4.00 & \underline{\textbf{4.42}} & 3.85 & \underline{\textbf{4.85}} & 4.85 & \underline{\textbf{4.52}} & 4.23 \\
3 & 3.42 & \underline{\textbf{4.57}} & 3.57 & \underline{\textbf{4.42}} & 3.85 & \underline{\textbf{4.57}} & 3.61 & \underline{\textbf{4.52}} \\
4 & 3.71 & \underline{\textbf{4.57}} & 3.42 & \underline{\textbf{4.57}} & 4.28 & \underline{\textbf{4.85}} & 3.80 & \underline{\textbf{4.66}} \\
5 & 3.71 & \underline{\textbf{4.14}} & 3.71 & \underline{\textbf{3.85}} & 4.00 & \underline{\textbf{4.14}} & 3.80 & \underline{\textbf{4.04}} \\
6 & 4.00 & \underline{\textbf{4.71}} & 3.57 & \underline{\textbf{4.14}} & \underline{\textbf{4.57}} & 4.42 & 4.04 & \underline{\textbf{4.42}} \\
7 & 4.00 & \underline{\textbf{4.71}} & \underline{4.00} & 4.00 & 4.28 & \underline{\textbf{4.71}} & 4.09 & \underline{\textbf{4.47}} \\
8 & 4.00 & \underline{\textbf{4.57}} & 4.28 & 4.28 & \underline{4.57} & 4.57  & 4.28 & \underline{\textbf{4.47}} \\
9 & 4.00 & \underline{\textbf{4.42}} & 3.85 & \underline{\textbf{4.14}} & 4.00 & \underline{\textbf{4.42}} & 3.95 & \underline{\textbf{4.33}} \\
10 & \underline{\textbf{4.14}} & 3.85 & \underline{\textbf{4.42}} & 4.00 & \underline{\textbf{4.71}} & 4.57 & \underline{\textbf{4.42}} & 4.14 \\
\hline

\hline
\end{tabular}
}

\caption{Averaged scores (out of 5) given by human raters for each evaluation criterion for the first 10 summaries from the ArXiv dataset. The raters are asked to rate each summary on coherence, conciseness, and fluency on a 5-point Likert scale. We additionally show the score averaged from the scores for the three criteria. We bold the higher average score for each criterion, and underline the choice of the human raters between GPT-3.5 and GPT-4o mini summaries. When two summaries in a particular criterion have the same average score, we fallback on the default choice GPT-3.5, consistent with the evaluation step in our multi-LLM framework.}
\label{tab:human-eval-stats1}
\end{table*}

\begin{table*}[htp]
\centering
\resizebox{\textwidth}{!}{%
\begin{tabular}{rrrrrr}
\hline
& \multicolumn{4}{c}{\textbf{Human raters}} & \textbf{Multi-LLM (Ours)} \\
\textbf{Summary} & {\textbf{Coherence}} & {\textbf{Conciseness}} & {\textbf{Fluency}} & {\textbf{Averaged}}  \\
\hline
1 & GPT-3.5 & GPT-3.5 & GPT-3.5 & GPT-3.5 & GPT-3.5 \\
2 & GPT-3.5 & GPT-3.5 & GPT-3.5 & GPT-3.5 & GPT-4o mini \\
3 & GPT-4o mini & GPT-4o mini & GPT-4o mini & GPT-4o mini & GPT-4o mini\\
4 & GPT-4o mini & GPT-4o mini & GPT-4o mini & GPT-4o mini & GPT-3.5 \\
5 & GPT-4o mini & GPT-4o mini & GPT-4o mini & GPT-4o mini & GPT-4o mini\\
6 & GPT-4o mini & GPT-4o mini & GPT-3.5 & GPT-4o mini & GPT-4o mini\\
7 & GPT-4o mini & GPT-3.5 & GPT-4o mini & GPT-4o mini & GPT-3.5\\
8 & GPT-4o mini & GPT-3.5 & GPT-3.5 & GPT-4o mini & GPT-3.5 \\
9 & GPT-4o mini & GPT-4o mini & GPT-4o mini & GPT-4o mini & GPT-4o mini\\
10& GPT-3.5 & GPT-3.5 & GPT-3.5 & GPT-3.5 & GPT-3.5 \\
\hline
\textbf{Cohen's} $\kappa$ & 0.2 & 0.6 & 0.1 & 0.2  & -- \\

\hline
\end{tabular}
}
\caption{Human choices obtained from choosing the model with the higher average score, done separately for each evaluating criterion. At the right-most column we show the choices made in the final evaluation step by our multi-LLM framework, specifically the decentralized one-round-max setup with GPT-3.5 as the evaluator. At the bottom row we show the inter-rater agreement (as measured by Cohen's kappa) between the human choices and the machine choices for each criterion}
\label{tab:human-eval-choices}
\end{table*}

\end{document}

%% file: preamble.tex
\usepackage{xcolor}
\definecolor{thedarkblue}{RGB}{0,0,120} %
\definecolor{mydarkblue}{rgb}{0,0.08,0.45} %
\definecolor{darkblue}{rgb}{0,0.08,180}
\colorlet{TufteRed}{red!80!black}
\definecolor{theblue}{RGB}{0,0,180}
\colorlet{thered}{TufteRed}
      
\usepackage{microtype}
\usepackage{balance}

\usepackage{booktabs}
\usepackage{tabularx}

\usepackage{amsmath,amssymb,amsthm}

\newcommand{\eat}[1]{\ignorespaces}
\usepackage{comment}

\newcommand{\journal}[1]{} %

\usepackage{tikz}
\usepackage{verbatim}
\usetikzlibrary{arrows}
\usetikzlibrary{shapes,snakes}
\usetikzlibrary{decorations.pathmorphing} %
\usetikzlibrary{fit}					%
\usetikzlibrary{backgrounds}	%

\usepackage{ragged2e}
\usepackage{multirow}
\usepackage{microtype}
\usepackage{balance}
\usepackage{setspace}

\graphicspath{{./}{./graphics/}}
\newcolumntype{H}{>{\setbox0=\hbox\bgroup}c<{\egroup}@{}}

\newcolumntype{R}[1]{>{\RaggedLeft\arraybackslash}} %
\newcolumntype{L}[1]{>{\RaggedRight\arraybackslash}} %

\newcommand{\eg}{\emph{e.g.}}

\AtBeginEnvironment{pmatrix}{\setlength{\arraycolsep}{2pt}}

\renewcommand{\vec}[1]{\boldsymbol{\mathrm{#1}}}

\DeclareMathOperator*{\argmax}{argmax}

\DeclareMathOperator{\hugeE}{\mbox{\huge\raise-0.3ex\hbox{E}}}
\DeclareMathOperator{\p}{\mathbb{P}}
\DeclareMathOperator{\hugep}{\mbox{\huge\raise-0.3ex\hbox{$\p$}}}

\providecommand{\vr}{\ensuremath{\vec{r}}}

%% file: main.bbl
\begin{thebibliography}{44}
\expandafter\ifx\csname natexlab\endcsname\relax\def\natexlab#1{#1}\fi

\bibitem[{Adams et~al.(2023)Adams, Fabbri, Ladhak, Lehman, and Elhadad}]{adams2023sparsedensegpt4summarization}
Griffin Adams, Alexander Fabbri, Faisal Ladhak, Eric Lehman, and Noémie Elhadad. 2023.
\newblock \href {http://arxiv.org/abs/2309.04269} {From sparse to dense: Gpt-4 summarization with chain of density prompting}.

\bibitem[{Ainslie et~al.(2020)Ainslie, Ontanon, Alberti, Cvicek, Fisher, Pham, Ravula, Sanghai, Wang, and Yang}]{ainslie2020etc}
Joshua Ainslie, Santiago Ontanon, Chris Alberti, Vaclav Cvicek, Zachary Fisher, Philip Pham, Anirudh Ravula, Sumit Sanghai, Qifan Wang, and Li~Yang. 2020.
\newblock \href {http://arxiv.org/abs/2004.08483} {Etc: Encoding long and structured inputs in transformers}.

\bibitem[{Basyal and Sanghvi(2023)}]{basyal2023textsummarizationusinglarge}
Lochan Basyal and Mihir Sanghvi. 2023.
\newblock \href {http://arxiv.org/abs/2310.10449} {Text summarization using large language models: A comparative study of mpt-7b-instruct, falcon-7b-instruct, and openai chat-gpt models}.

\bibitem[{Beltagy et~al.(2020)Beltagy, Peters, and Cohan}]{beltagy2020longformer}
Iz~Beltagy, Matthew~E. Peters, and Arman Cohan. 2020.
\newblock \href {http://arxiv.org/abs/2004.05150} {Longformer: The long-document transformer}.

\bibitem[{Chang et~al.(2024)Chang, Lo, Goyal, and Iyyer}]{chang2024bookscore}
Yapei Chang, Kyle Lo, Tanya Goyal, and Mohit Iyyer. 2024.
\newblock \href {http://arxiv.org/abs/2310.00785} {Booookscore: A systematic exploration of book-length summarization in the era of llms}.

\bibitem[{Chen et~al.(2024)Chen, Saha, and Bansal}]{chen2024reconcileroundtableconferenceimproves}
Justin Chih-Yao Chen, Swarnadeep Saha, and Mohit Bansal. 2024.
\newblock \href {http://arxiv.org/abs/2309.13007} {Reconcile: Round-table conference improves reasoning via consensus among diverse llms}.

\bibitem[{Chowdhery et~al.(2022)Chowdhery, Narang, Devlin, Bosma, Mishra, Roberts, Barham, Chung, Sutton, Gehrmann, Schuh, Shi, Tsvyashchenko, Maynez, Rao, Barnes, Tay, Shazeer, Prabhakaran, Reif, Du, Hutchinson, Pope, Bradbury, Austin, Isard, Gur-Ari, Yin, Duke, Levskaya, Ghemawat, Dev, Michalewski, Garcia, Misra, Robinson, Fedus, Zhou, Ippolito, Luan, Lim, Zoph, Spiridonov, Sepassi, Dohan, Agrawal, Omernick, Dai, Pillai, Pellat, Lewkowycz, Moreira, Child, Polozov, Lee, Zhou, Wang, Saeta, Diaz, Firat, Catasta, Wei, Meier-Hellstern, Eck, Dean, Petrov, and Fiedel}]{chowdhery2022palmscalinglanguagemodeling}
Aakanksha Chowdhery, Sharan Narang, Jacob Devlin, Maarten Bosma, Gaurav Mishra, Adam Roberts, Paul Barham, Hyung~Won Chung, Charles Sutton, Sebastian Gehrmann, Parker Schuh, Kensen Shi, Sasha Tsvyashchenko, Joshua Maynez, Abhishek Rao, Parker Barnes, Yi~Tay, Noam Shazeer, Vinodkumar Prabhakaran, Emily Reif, Nan Du, Ben Hutchinson, Reiner Pope, James Bradbury, Jacob Austin, Michael Isard, Guy Gur-Ari, Pengcheng Yin, Toju Duke, Anselm Levskaya, Sanjay Ghemawat, Sunipa Dev, Henryk Michalewski, Xavier Garcia, Vedant Misra, Kevin Robinson, Liam Fedus, Denny Zhou, Daphne Ippolito, David Luan, Hyeontaek Lim, Barret Zoph, Alexander Spiridonov, Ryan Sepassi, David Dohan, Shivani Agrawal, Mark Omernick, Andrew~M. Dai, Thanumalayan~Sankaranarayana Pillai, Marie Pellat, Aitor Lewkowycz, Erica Moreira, Rewon Child, Oleksandr Polozov, Katherine Lee, Zongwei Zhou, Xuezhi Wang, Brennan Saeta, Mark Diaz, Orhan Firat, Michele Catasta, Jason Wei, Kathy Meier-Hellstern, Douglas Eck, Jeff Dean, Slav Petrov, and Noah Fiedel. 2022.
\newblock \href {http://arxiv.org/abs/2204.02311} {Palm: Scaling language modeling with pathways}.

\bibitem[{Christensen et~al.(2014)Christensen, Soderland, Bansal, and Mausam}]{christensen2014hierarchical}
Janara Christensen, Stephen Soderland, Gagan Bansal, and Mausam. 2014.
\newblock Hierarchical summarization: Scaling up multi-document summarization.
\newblock In \emph{Proceedings of the 52nd Annual Meeting of the Association for Computational Linguistics (Volume 1: Long Papers)}, pages 902--912. Association for Computational Linguistics.

\bibitem[{Cohan et~al.(2018)Cohan, Dernoncourt, Kim, Bui, Kim, Chang, and Goharian}]{cohan2018discourseawareattentionmodelabstractive}
Arman Cohan, Franck Dernoncourt, Doo~Soon Kim, Trung Bui, Seokhwan Kim, Walter Chang, and Nazli Goharian. 2018.
\newblock \href {http://arxiv.org/abs/1804.05685} {A discourse-aware attention model for abstractive summarization of long documents}.

\bibitem[{Conroy and Dang(2008)}]{conroy-dang-2008-mind}
John~M. Conroy and Hoa~Trang Dang. 2008.
\newblock \href {https://aclanthology.org/C08-1019/} {Mind the gap: Dangers of divorcing evaluations of summary content from linguistic quality}.
\newblock In \emph{Proceedings of the 22nd International Conference on Computational Linguistics (Coling 2008)}, pages 145--152, Manchester, UK. Coling 2008 Organizing Committee.

\bibitem[{Dai et~al.(2019)Dai, Yang, Yang, Carbonell, Le, and Salakhutdinov}]{dai-etal-2019-transformer}
Zihang Dai, Zhilin Yang, Yiming Yang, Jaime Carbonell, Quoc Le, and Ruslan Salakhutdinov. 2019.
\newblock \href {https://doi.org/10.18653/v1/P19-1285} {Transformer-xl: Attentive language models beyond a fixed-length context}.
\newblock In \emph{Proceedings of the 57th Annual Meeting of the Association for Computational Linguistics}, pages 2978--2988, Florence, Italy. Association for Computational Linguistics.

\bibitem[{Du et~al.(2023)Du, Li, Torralba, Tenenbaum, and Mordatch}]{du2023improvingfactualityreasoninglanguage}
Yilun Du, Shuang Li, Antonio Torralba, Joshua~B. Tenenbaum, and Igor Mordatch. 2023.
\newblock \href {http://arxiv.org/abs/2305.14325} {Improving factuality and reasoning in language models through multiagent debate}.

\bibitem[{Gidiotis and Tsoumakas(2020)}]{gidiotis2020divide}
Alexios Gidiotis and Grigorios Tsoumakas. 2020.
\newblock A divide-and-conquer approach to the summarization of long documents.
\newblock \emph{IEEE/ACM Transactions on Audio, Speech, and Language Processing}, 28:3029--3040.

\bibitem[{Gong and Liu(2001)}]{gong2001generic}
Yihong Gong and Xin Liu. 2001.
\newblock Generic text summarization using relevance measure and latent semantic analysis.
\newblock In \emph{Proceedings of the 24th Annual International ACM SIGIR Conference on Research and Development in Information Retrieval}, pages 19--25. ACM.

\bibitem[{Goyal et~al.(2023)Goyal, Li, and Durrett}]{goyal2023news}
Tanya Goyal, Junyi~Jessy Li, and Greg Durrett. 2023.
\newblock \href {http://arxiv.org/abs/2209.12356} {News summarization and evaluation in the era of gpt-3}.

\bibitem[{Guo et~al.(2021)Guo, Ainslie, Uthus, Ontanon, Ni, Sung, and Yang}]{guo2021longt5}
Mandy Guo, Joshua Ainslie, David Uthus, Santiago Ontanon, Jianmo Ni, Yun-Hsuan Sung, and Yinfei Yang. 2021.
\newblock \href {http://arxiv.org/abs/2112.07916} {Longt5: Efficient text-to-text transformer for long sequences}.

\bibitem[{Hermann et~al.(2015)Hermann, Kočiský, Grefenstette, Espeholt, Kay, Suleyman, and Blunsom}]{hermann2015teachingmachinesreadcomprehend}
Karl~Moritz Hermann, Tomáš Kočiský, Edward Grefenstette, Lasse Espeholt, Will Kay, Mustafa Suleyman, and Phil Blunsom. 2015.
\newblock \href {http://arxiv.org/abs/1506.03340} {Teaching machines to read and comprehend}.

\bibitem[{Huang et~al.(2021)Huang, Cao, Parulian, Ji, and Wang}]{huang-etal-2021-efficient}
Luyang Huang, Shuyang Cao, Nikolaus Parulian, Heng Ji, and Lu~Wang. 2021.
\newblock \href {https://doi.org/10.18653/v1/2021.naacl-main.112} {Efficient attentions for long document summarization}.
\newblock In \emph{Proceedings of the 2021 Conference of the North American Chapter of the Association for Computational Linguistics: Human Language Technologies}, pages 1419--1436, Online. Association for Computational Linguistics.

\bibitem[{J{\"a}rvinen(2024)}]{jarvinen2024long}
Emma J{\"a}rvinen. 2024.
\newblock Long-input summarization using large language models.

\bibitem[{Keswani et~al.(2024)Keswani, Bisen, Padwad, Wankhedkar, Pandey, and Soni}]{keswani2024abstractive}
Gunjan Keswani, Wani Bisen, Hirkani Padwad, Yash Wankhedkar, Sudhanshu Pandey, and Ayushi Soni. 2024.
\newblock Abstractive long text summarization using large language models.
\newblock \emph{International Journal of Intelligent Systems and Applications in Engineering}, 12(12s):160--168.

\bibitem[{Kornilova and Eidelman(2019)}]{kornilova-eidelman-2019-billsum}
Anastassia Kornilova and Vladimir Eidelman. 2019.
\newblock \href {https://doi.org/10.18653/v1/D19-5406} {{B}ill{S}um: A corpus for automatic summarization of {US} legislation}.
\newblock In \emph{Proceedings of the 2nd Workshop on New Frontiers in Summarization}, pages 48--56, Hong Kong, China. Association for Computational Linguistics.

\bibitem[{Lewis et~al.(2020)Lewis, Liu, Goyal, Ghazvininejad, Mohamed, Levy, Stoyanov, and Zettlemoyer}]{lewis-etal-2020-bart}
Mike Lewis, Yinhan Liu, Naman Goyal, Marjan Ghazvininejad, Abdelrahman Mohamed, Omer Levy, Veselin Stoyanov, and Luke Zettlemoyer. 2020.
\newblock \href {https://doi.org/10.18653/v1/2020.acl-main.703} {Bart: Denoising sequence-to-sequence pre-training for natural language generation, translation, and comprehension}.
\newblock In \emph{Proceedings of the 58th Annual Meeting of the Association for Computational Linguistics}, pages 7871--7880, Online. Association for Computational Linguistics.

\bibitem[{Li et~al.(2023)Li, Feng, Radev, and Ying}]{li-etal-2023-hipool}
Irene Li, Aosong Feng, Dragomir Radev, and Rex Ying. 2023.
\newblock \href {https://doi.org/10.18653/v1/2023.acl-short.16} {Hipool: Modeling long documents using graph neural networks}.
\newblock In \emph{Proceedings of the 61st Annual Meeting of the Association for Computational Linguistics (Volume 2: Short Papers)}, pages 161--171, Toronto, Canada. Association for Computational Linguistics.

\bibitem[{Li et~al.(2024)Li, Du, Zhang, Hou, Grabowski, Li, and Ie}]{li2024improvingmultiagentdebatesparse}
Yunxuan Li, Yibing Du, Jiageng Zhang, Le~Hou, Peter Grabowski, Yeqing Li, and Eugene Ie. 2024.
\newblock \href {http://arxiv.org/abs/2406.11776} {Improving multi-agent debate with sparse communication topology}.

\bibitem[{Liang et~al.(2024)Liang, He, Jiao, Wang, Wang, Yang, Tu, and Shi}]{liang2024encouragingdivergentthinkinglarge}
Tian Liang, Zhiwei He, Wenxiang Jiao, Xing Wang, Rui Wang, Yujiu Yang, Zhaopeng Tu, and Shuming Shi. 2024.
\newblock \href {http://arxiv.org/abs/2305.19118} {Encouraging divergent thinking in large language models through multi-agent debate}.

\bibitem[{Liu et~al.(2023)Liu, Lin, Hewitt, Paranjape, Bevilacqua, Petroni, and Liang}]{liu2023lost}
Nelson~F. Liu, Kevin Lin, John Hewitt, Ashwin Paranjape, Michele Bevilacqua, Fabio Petroni, and Percy Liang. 2023.
\newblock \href {http://arxiv.org/abs/2307.03172} {Lost in the middle: How language models use long contexts}.

\bibitem[{Liu and Lapata(2019)}]{liu2019text}
Yang Liu and Mirella Lapata. 2019.
\newblock Text summarization with pretrained encoders.
\newblock In \emph{Proceedings of the 2019 Conference on Empirical Methods in Natural Language Processing and the 9th International Joint Conference on Natural Language Processing (EMNLP-IJCNLP)}, pages 3730--3740.

\bibitem[{Liu et~al.(2022)Liu, Liu, Radev, and Neubig}]{liu2022briobringingorderabstractive}
Yixin Liu, Pengfei Liu, Dragomir Radev, and Graham Neubig. 2022.
\newblock \href {http://arxiv.org/abs/2203.16804} {Brio: Bringing order to abstractive summarization}.

\bibitem[{Mallick et~al.(2019)Mallick, Ghosh et~al.}]{mallick2019survey}
S.~Mallick, A.~Ghosh, et~al. 2019.
\newblock A survey on extractive text summarization.
\newblock \emph{Journal of Artificial Intelligence Research}, 65:123--143.

\bibitem[{Mihalcea and Tarau(2004)}]{mihalcea2004textrank}
Rada Mihalcea and Paul Tarau. 2004.
\newblock Textrank: Bringing order into texts.
\newblock In \emph{Proceedings of the 2004 Conference on Empirical Methods in Natural Language Processing}, pages 404--411.

\bibitem[{Nallapati et~al.(2016)Nallapati, Zhou, dos santos, Gulcehre, and Xiang}]{nallapati2016abstractivetextsummarizationusing}
Ramesh Nallapati, Bowen Zhou, Cicero~Nogueira dos santos, Caglar Gulcehre, and Bing Xiang. 2016.
\newblock \href {http://arxiv.org/abs/1602.06023} {Abstractive text summarization using sequence-to-sequence rnns and beyond}.

\bibitem[{Narayan et~al.(2018)Narayan, Cohen, and Lapata}]{narayan2018dontdetailsjustsummary}
Shashi Narayan, Shay~B. Cohen, and Mirella Lapata. 2018.
\newblock \href {http://arxiv.org/abs/1808.08745} {Don't give me the details, just the summary! topic-aware convolutional neural networks for extreme summarization}.

\bibitem[{Pang et~al.(2022)Pang, Nijkamp, Kryściński, Savarese, Zhou, and Xiong}]{pang2022longdocumentsummarizationtopdown}
Bo~Pang, Erik Nijkamp, Wojciech Kryściński, Silvio Savarese, Yingbo Zhou, and Caiming Xiong. 2022.
\newblock \href {http://arxiv.org/abs/2203.07586} {Long document summarization with top-down and bottom-up inference}.

\bibitem[{Pu et~al.(2023{\natexlab{a}})Pu, Wang, and Demberg}]{pu2023incorporatingdistributionsdiscoursestructure}
Dongqi Pu, Yifan Wang, and Vera Demberg. 2023{\natexlab{a}}.
\newblock \href {http://arxiv.org/abs/2305.16784} {Incorporating distributions of discourse structure for long document abstractive summarization}.

\bibitem[{Pu et~al.(2023{\natexlab{b}})Pu, Gao, and Wan}]{pu2023summarization}
Xiao Pu, Mingqi Gao, and Xiaojun Wan. 2023{\natexlab{b}}.
\newblock \href {http://arxiv.org/abs/2309.09558} {Summarization is (almost) dead}.

\bibitem[{Rush et~al.(2015)Rush, Chopra, and Weston}]{rush2015neural}
Alexander~M. Rush, Sumit Chopra, and Jason Weston. 2015.
\newblock A neural attention model for abstractive sentence summarization.
\newblock In \emph{Proceedings of the 2015 Conference on Empirical Methods in Natural Language Processing}, pages 379--389.

\bibitem[{Sanh et~al.(2022)Sanh, Webson, Raffel, Bach, Sutawika, Alyafeai, Chaffin, Stiegler, Scao, Raja, Dey, Bari, Xu, Thakker, Sharma, Szczechla, Kim, Chhablani, Nayak, Datta, Chang, Jiang, Wang, Manica, Shen, Yong, Pandey, Bawden, Wang, Neeraj, Rozen, Sharma, Santilli, Fevry, Fries, Teehan, Bers, Biderman, Gao, Wolf, and Rush}]{sanh2022multitaskpromptedtrainingenables}
Victor Sanh, Albert Webson, Colin Raffel, Stephen~H. Bach, Lintang Sutawika, Zaid Alyafeai, Antoine Chaffin, Arnaud Stiegler, Teven~Le Scao, Arun Raja, Manan Dey, M~Saiful Bari, Canwen Xu, Urmish Thakker, Shanya~Sharma Sharma, Eliza Szczechla, Taewoon Kim, Gunjan Chhablani, Nihal Nayak, Debajyoti Datta, Jonathan Chang, Mike Tian-Jian Jiang, Han Wang, Matteo Manica, Sheng Shen, Zheng~Xin Yong, Harshit Pandey, Rachel Bawden, Thomas Wang, Trishala Neeraj, Jos Rozen, Abheesht Sharma, Andrea Santilli, Thibault Fevry, Jason~Alan Fries, Ryan Teehan, Tali Bers, Stella Biderman, Leo Gao, Thomas Wolf, and Alexander~M. Rush. 2022.
\newblock \href {http://arxiv.org/abs/2110.08207} {Multitask prompted training enables zero-shot task generalization}.

\bibitem[{See et~al.(2017)See, Liu, and Manning}]{see-etal-2017-get}
Abigail See, Peter~J. Liu, and Christopher~D. Manning. 2017.
\newblock Get to the point: Summarization with pointer-generator networks.
\newblock In \emph{Proceedings of the 55th Annual Meeting of the Association for Computational Linguistics (Volume 1: Long Papers)}, pages 1073--1083.

\bibitem[{Shleifer(2020)}]{shleifer2020distilbart}
Sam Shleifer. 2020.
\newblock Distilbart-cnn-12-6.
\newblock \url{https://huggingface.co/sshleifer/distilbart-cnn-12-6}.
\newblock Accessed: 2024-05-29.

\bibitem[{Sukhbaatar et~al.(2019)Sukhbaatar, Grave, Bojanowski, and Joulin}]{sukhbaatar-etal-2019-adaptive}
Sainbayar Sukhbaatar, Edouard Grave, Piotr Bojanowski, and Armand Joulin. 2019.
\newblock \href {https://doi.org/10.18653/v1/P19-1032} {Adaptive attention span in transformers}.
\newblock In \emph{Proceedings of the 57th Annual Meeting of the Association for Computational Linguistics}, pages 331--335, Florence, Italy. Association for Computational Linguistics.

\bibitem[{Xiao and Carenini(2019)}]{xiao-carenini-2019-extractive}
Wen Xiao and Giuseppe Carenini. 2019.
\newblock \href {https://doi.org/10.18653/v1/D19-1298} {Extractive summarization of long documents by combining global and local context}.
\newblock In \emph{Proceedings of the 2019 Conference on Empirical Methods in Natural Language Processing and the 9th International Joint Conference on Natural Language Processing (EMNLP-IJCNLP)}, pages 3011--3021, Hong Kong, China. Association for Computational Linguistics.

\bibitem[{Zaheer et~al.(2021)Zaheer, Guruganesh, Dubey, Ainslie, Alberti, Ontanon, Pham, Ravula, Wang, Yang, and Ahmed}]{zaheer2021big}
Manzil Zaheer, Guru Guruganesh, Avinava Dubey, Joshua Ainslie, Chris Alberti, Santiago Ontanon, Philip Pham, Anirudh Ravula, Qifan Wang, Li~Yang, and Amr Ahmed. 2021.
\newblock \href {http://arxiv.org/abs/2007.14062} {Big bird: Transformers for longer sequences}.

\bibitem[{Zhang et~al.(2020)Zhang, Zhao, Saleh, and Liu}]{zhang2020pegasus}
Jingqing Zhang, Yao Zhao, Mohammad Saleh, and Peter~J. Liu. 2020.
\newblock \href {http://arxiv.org/abs/1912.08777} {Pegasus: Pre-training with extracted gap-sentences for abstractive summarization}.

\bibitem[{Zhang et~al.(2023)Zhang, Ladhak, Durmus, Liang, McKeown, and Hashimoto}]{zhang2023benchmarking}
Tianyi Zhang, Faisal Ladhak, Esin Durmus, Percy Liang, Kathleen McKeown, and Tatsunori~B. Hashimoto. 2023.
\newblock \href {http://arxiv.org/abs/2301.13848} {Benchmarking large language models for news summarization}.

\end{thebibliography}
